\documentclass[letterpaper]{article} 
\usepackage{aaai24}  
\usepackage{times}  
\usepackage{helvet}  
\usepackage{courier}  
\usepackage[hyphens]{url}  
\usepackage{graphicx} 
\usepackage{natbib}  
\usepackage{caption} 
\usepackage{algorithm}
\usepackage{algorithmic}
\usepackage{tikz}
\usetikzlibrary{positioning}
\usetikzlibrary{calc}
\usetikzlibrary{arrows,shapes,backgrounds}
\usepackage{newfloat}
\usepackage{listings}
\usepackage{booktabs}
\usepackage[shortlabels]{enumitem}
\DeclareCaptionStyle{ruled}{labelfont=normalfont,labelsep=colon,strut=off} 
\floatstyle{ruled}
\newfloat{listing}{tb}{lst}{}
\floatname{listing}{Listing}
\title{A Theory of Non-Acyclic Generative Flow Networks}
\author {
    Leo Maxime Brunswic\textsuperscript{\rm 1}\textsuperscript{\rm 2},
    Yinchuan Li\textsuperscript{\rm 3*},
    Yushun Xu\textsuperscript{\rm 1},
    Shangling Jui\textsuperscript{\rm 1},
    Lizhuang Ma\textsuperscript{\rm 2}
}
\affiliations {
    \textsuperscript{\rm 1} Huawei Shanghai Research Center\\
    \textsuperscript{\rm 2} Jiaotong Shanghai University\\
    \textsuperscript{\rm 3} Huawei Noah's Ark Lab, Beijing, China\\
    \{leomaximebrunswic, liyinchuan, jui.shangling\}@huawei.com, xuyushun1@hisilicon.com, ma-lz@cs.sjtu.edu.cn
}

\usepackage{amsmath,amssymb,MnSymbol,amsthm}
\usepackage[all,cmtip,color]{xy}

\def\NN{\mathbb{N}}
\def\Z{\mathbb{Z}}

\def\Init{{s_0}}
\def\Final{{s_f}}

\def\RR{\mathbb{R}}

\def\d{\mathrm{d}}

\def\Fin{F_{\mathrm{in}}}
\def\Fout{F_{\mathrm{out}}}
\def\Finit{F_{\mathrm{init}}}
\def\Foutbar{\overline{F}_{\mathrm{out}}}

\def\ForwardPolicy{\pi_f}
\def\BackwardPolicy{\pi_b}
\def\LossFM{\mathcal L_{\text{FM}}}
\def\LossDB{\mathcal L_{\text{DB}}}
\def\LossTB{\mathcal L_{\text{TB}}}
\def\Div{\mathrm{div}}
\definecolor{gray1}{gray}{0.8}
\definecolor{gray2}{gray}{0.9}

\newtheorem*{rep@theorem}{\rep@title}
\newcommand{\newreptheorem}[2]{%
\newenvironment{rep#1}[1]{%
 \def\rep@title{#2 \ref{##1}}%
 \begin{rep@theorem}}%
 {\end{rep@theorem}}}
\makeatother

\newtheorem{defi}{Definition}
\newtheorem{ex}{Example}[defi]
\newtheorem{cor}{Corollary}

\newtheorem{prop}{Proposition}

\newtheorem{theo}{Theorem}
\newtheorem{example}{Example}

\newtheorem{lemma}{Lemma}
\newreptheorem{lemma}{Lemma}
\newreptheorem{theorem}{Theorem}
\newreptheorem{prop}{Proposition}

\begin{document}

\maketitle
\begin{abstract}
 GFlowNets is a novel flow-based method for learning a stochastic policy to generate objects via a sequence of actions and with probability proportional to a given positive reward.
We contribute to relaxing hypotheses limiting the application range of GFlowNets, in particular: acyclicity (or lack thereof).
To this end, we extend the theory of GFlowNets on measurable spaces which includes continuous state spaces without cycle restrictions, and provide a generalization of cycles in this generalized context.
We show that losses used so far push flows to get stuck into cycles and we define  a family of losses solving this issue. Experiments on graphs and continuous tasks validate those principles.
\end{abstract}

\section{Introduction}\label{sec:GFLOW}

Bengio et al. \cite{bengio2021flow,bengio2021gflownet,zhang2022unifying,madan2022learning,malkin2022trajectory} introduced GFlowNets (GFN): a new method for diverse candidate generation. The objective of GFN is to sample states of a Directed Acyclic Graph (DAG) proportionally to some reward function (see Section \ref{sec:DAG_GFN} for quick introduction).
In this way, GFN may be compared to MCMC \cite{brooks2011handbook} and distributional reinforcement learning methods \cite{bellemare2017distributional}.  Experiments conducted by Bengio et al. on challenging tasks compared GFN favorably to MCMC-based methods. GFlowNets has also been successfully used in GNN \cite{li2023dag,li2023generative}, domain adaptation \cite{zhu2023generalized}, optimization \cite{zhang2023let}, large language model training \cite{LiLSH23}, offline data \cite{WangSHL23} and other fields.


Restricting GFN to DAG was initially motivated by specific applications in which solutions are built by applying sequences of constructors such as in the case of molecule generation. This seems however too restrictive: both GFN and MCMC methods sample an unnormalized distribution, it is tempting to try using GFN as a replacement for MCMC altogether (raytracing \cite{veach1997metropolis} is a typical industrial application of MCMC) leading to the need for a generalization of GFN to continuous state spaces. Moreover, some tasks have intrinsic symmetries that are better modeled by a non-acyclic graph, say a Cayley graph like the one of  Rubick's cube. Worse, state space and transitions may be constrained in such a way that cycles are unavoidable: one may think of a game in which the adversary may force to return to a previous state. Recent work \cite{li2023cflownets,lahlou2023theory} made attempts at the former but the latter is still unaddressed.

\begin{figure}[h]
\centering
\begin{centering}
\includegraphics[width=0.45\linewidth]{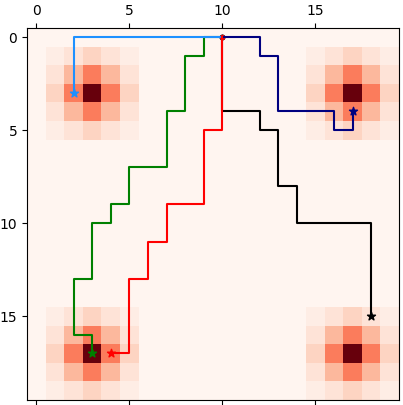}\includegraphics[width=0.45\linewidth]{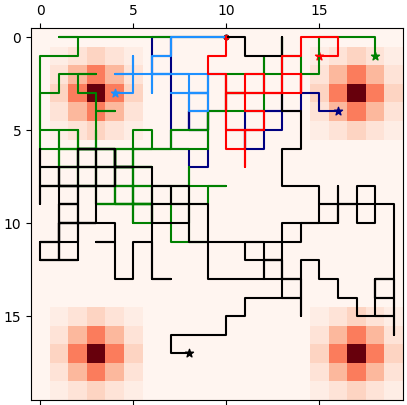}
\end{centering}
\caption{Paths generated on a grid using: \textbf{Left:} non-acyclic loss (ours); \textbf{Right:} FM loss (\cite{bengio2021flow}).
\label{fig:HGpathsintro}}
\end{figure}

The main aim of this paper is to clarify the issues brought up during the training of generative flows on spaces with cycles. In particular, we study their effects on gradient descent minimizations of the so-called Flow-Matching, Detailed Balance, and Trajectory Balance losses introduced in the previous aforementioned works to train GFN.

Considering the level of generalization of the present work is a few steps above that of the initial GFN framework, we use the term ``Generative Flows" so that GFN correspond to the special case of Generative Flows on DAG.

\subsubsection{Main Contributions \& Outline of the Paper}1) We describe in section \ref{sec:stabilitygraph} a property of losses on graphs leading flows to be trapped into loops and define a notion of stability in order to control this behavior which is detrimental to training and increases inference cost. 
2) Section \ref{sec:theoretic} describes the elementary measure theoretical generalization of GFN slightly differently from that of \citet{lahlou2023theory} lifting the built-in acyclicity constraint.  The mathematical framework is enriched in particular with the definition of a suitable generalization of cycles. 3) In section \ref{sec:manage_cycles}, we reformulate Flow Matching, Detailed Balance, and Trajectory Balance losses as variations of $f$-divergences and show that such losses are unstable in the sense we introduced. We propose a family of stable losses and regularizations to increase stability. 4)  We demonstrate in section \ref{sec:experiments} the effect of stability by solving toy problems on Hypergrids, Cayley graphs (GFN), and a continuous task from Gym \cite {brockman2016openai,towers_gymnasium_2023} (CFlowNets). Appendix A and B respectively provide extra theoretical results and proofs.

\section{Background: Generative Flows on DAG} \label{sec:DAG_GFN}
Following \citet{bengio2021flow} and \citet{bengio2021gflownet},
let  $G$ be a {\it directed acyclic graph} (DAG) given by its (countable) {\it state set} $S:=S^*\cup\{\Init, \Final\}$ and its  {\it edges} set $E$\footnote{Our formulation differs slightly from that of  \cite{bengio2021flow} in which `edges' are referred to as `actions'.}  with source and sink states $s_0$ and  $s_f$ \footnote{A source is a state having no ongoing edge, a sink is a state having no outgoing edge.}
For the sake of simplicity, we assume $G$ is not a multigraph. An {\it edgeflow} on $G$ is an assignment $F$ of a non-negative value on each edge, $\forall s\rightarrow s', F(s\rightarrow s')\geq 0$.
\begin{defi}
A flow is an edgeflow $F$ satisfying the so-called flow-matching constraint (\ref{equ:FMconst_graph}). Given a reward $R:S^* \rightarrow \RR_+$, an edgeflow is a $R$-edgeflow if it satisfies the reward constraint (\ref{equ:Rewardconst_graph}). A $R$-flow is an edgeflow satisfying both constraints, as the following
\begin{eqnarray} \label{equ:FMconst_graph}
 \forall s\in S^*,&&\underbrace{\sum_{s'\rightarrow s} F(s'\rightarrow s)}_{\text{Ingoing Flow}} =  \underbrace{\sum_{s\rightarrow s'} F(s\rightarrow s')}_{\text{Outgoing Flow}} \\
  \forall s\in S^*,&&F(s\rightarrow s_f) = R(s).\label{equ:Rewardconst_graph}
\end{eqnarray}
\end{defi}

 Any edgeflow on such a finite graph $G$ induces a Markov chain $(s_0,s_1,s_2,\cdots,s_f)$ starting at the source $\Init$, stopping at the sink $s_f$ and
at each time $t$ such that $s_t\neq \Final$:
\begin{equation}\label{equ:transition_graph}
\mathbb P(s_{t+1}  = s| s_t) = \frac{F(s_t\rightarrow
s)}{\sum_{s_t\rightarrow s'} F(s_t\rightarrow s') } 
\end{equation}
whenever this formula makes sense. This Markov chain samples a state $s$ at time $\tau$ if $s_\tau=s$ and $s_{\tau+1} = \Final$, the time $\tau$ is the {\it sampling time}.
The fundamental result of \citet{bengio2021flow} is that equations \eqref{equ:FMconst_graph}-\eqref{equ:transition_graph} imply that at the sampling time, we have $s_\tau$ is distributed proportionally to the reward.

One trains a parameterized model of edgeflows to satisfy the flow-matching constraint while the reward constraint is enforced by the implementation.  To train such a model via gradient descent, one thus defines a loss $\mathcal L$ that is minimized if and only if the flow-matching constraint (\ref{equ:FMconst_graph}) is satisfied.
So far, three families of losses have been proposed: Flow-Matching loss (FM) \cite{bengio2021flow}, Detailed Balance loss (DB) \cite{bengio2021gflownet}   and Trajectory Balance loss (TB) \cite{malkin2022trajectory}, respectively as the following:
\begin{align}
\label{FM-loss}
    \LossFM(F^\theta) &= \mathbb E \sum_{t=1}^{\tau} \log ^2\frac{\sum_{s_t\rightarrow s'} F^{\theta}(s_t\rightarrow s')} {\sum_{s'\rightarrow s_t} F^{\theta}(s'\rightarrow s_t)}, \\
    \label{DB-loss}
    \LossDB(F_f^\theta,F_b^\theta) &= \mathbb E \sum_{t=0}^{\tau-1} \log ^2\frac{F_{\mathrm{out}}^{\theta,f}(s_t)\pi_f(s_t\rightarrow s_{t+1})} {F_{\mathrm{out}}^{\theta ,f}(s_{t+1})\pi_b(s_{t}\rightarrow s_{t+1}) }, \\
    \label{TB-loss}
    \LossTB(F_f^\theta,F_b^\theta) &= \mathbb E  \log ^2\frac{F_{\mathrm{out}}^{\theta,f}(s_0)\prod_{t=0}^{\tau}\pi_f(s_t\rightarrow s_{t+1})} {R(s_\tau)\prod_{t=0}^{\tau-1} \pi_b(s_t\rightarrow s_{t+1}) }.
\end{align}
The expectations above are taken over a distribution of paths $\Init\rightarrow s_1\rightarrow \cdots \rightarrow s_\tau\rightarrow s_f$   
and we used  $\ForwardPolicy(s\rightarrow s'):= \frac{F_f(s_t\rightarrow
s)}{\sum_{s_t\rightarrow s'} F_f(s_t\rightarrow s') }$ and $\BackwardPolicy(s\rightarrow s'):= \mathbb P(s_t = s | s_{t+1}=s')=\frac{F_b(s\rightarrow s')}{\sum_{s''\rightarrow s'} F_b(s''\rightarrow s')}$.
Obvious variations may be built by replacing $x,y\mapsto \log^2(\frac{x}{y})$ by any distance-like function.

\section{From DAG to Measurable Spaces}
\label{sec:DAGtomeasurable}
\subsection{Stability, Cycles and 0-Flows on Graphs}
\label{sec:stabilitygraph}
\subsubsection{Loss Instability on Non-Acyclic Graphs}
All definitions considered in the case of DAG still make sense on Directed Graphs without the acyclicity assumption. Bengio's sampling theorem is still true and will be derived as a subcase of a more general statement given on measurable state spaces.
However, training GFlowNets fails in the presence of cycles: one can see that training converges toward infinite edgeflow along cycles. In particular, the expected sampling time $\tau$ grows to infinity: $\mathbb E(\tau)\rightarrow +\infty$.
For example, consider the following families of edgeflows $F^{\theta}$ parameterized by a free non-negative parameter $\theta:=(f_1,f_2,f_3,c)$.
$$
\xymatrix{
 \Init  \ar[r]^{f_1}   &A  \ar[r]^{f_2}  &B \ar@/_1pc/[r]^{f_3+c} &C  \ar@/_1pc/[l]^{c}\ar[r]^1&\Final
}.
$$
Flows on this graph satisfying the reward constraints are exactly the $F^\theta$ corresponding to $c\geq 0$ and $f_1=f_2=f_3=1$. However, $\LossFM$ may be minimized by letting $c\rightarrow +\infty$ and $f_1,f_2,f_3\rightarrow 0$.
This is captioned by $\partial_c \LossFM(F^\theta)<0$, this problem also occurs for $\LossDB$ and  $\LossTB$.

\subsubsection{Linearity}
A typical implementation of a flow on a graph will be via an MLP with two heads: a softmax and a scalar. They respectively output the forward policy and the outgoing flow at a given state. From an analytical view point, however, edgeflows on graphs are non-negative real-valued maps on the edges set: the space of edgeflows is thus a subdomain of a vector space.  Furthermore, the reward and flowmatching constraints are linear, we deduce that sums and convex combinations of edgeflows as well as directional derivatives of a functional on edgeflows in the direction of some edgeflow makes sense.  Precise description of the space of flows is given in appendix.
\subsubsection{0-Flows}
Cycles induce flows on graphs in a natural way: for a cycle $\gamma=(s_1\rightarrow\cdots \rightarrow s_\ell\rightarrow s_1)$, we may define the $0$-flow $1_\gamma$ by  $1_\gamma(s\rightarrow s')= 1$ if the edge $s\rightarrow s'$ is in the cycle $\gamma$ and $0$ otherwise. Such a $1_\gamma$ is a flow but also satisfies the reward constraint for null reward: it is a $0$-flow. 
\begin{defi}\label{def-0-flows} A $0$-flow is a flow satisfying the reward constraint (\ref{equ:Rewardconst_graph}) for the null reward $R=0$.
\end{defi}
 $0$-flows are a nice way to speak about cycles as edgeflows: taking the derivative of a functional on edgeflows in the direction of a cycle makes sense.
 Moreover, on graphs, the space of $0$-flows is generated by cycles (see appendix).  
\subsubsection{Formalization of Stability}
Next, define an edgeflow $F_0$ as a subflow of an edgeflow $F$ with $F_0\leq F$ as functions defined on edges. We can have the following stability definition.
\begin{defi}[Stability]

A loss $\mathcal L$ acting on families of flows $(F_1,\cdots, F_p)$ is stable if adding a $0$-flow does not decrease the loss. More formally:
if for any $0$-flow $F_0$
$$ \mathcal L(F_1+F_0,\cdots,F_p+F_0) \geq \mathcal L(F_1,\cdots,F_p),$$
then the loss is stable.
\end{defi}

\begin{lemma}
   A sufficient condition for stability is 
    that $$ \partial _{F_0} \mathcal L(F_1,\cdots,F_p) \geq 0$$
    for all $0$-flow  $F_0$ which is a subflow of each $F_i$.
\end{lemma}
\subsubsection{Stabilizing Regularization}
The first consequence of the discussion above is the use of stabilizing regularizations. 

\begin{theo} \label{theo:cv}   Let
\begin{equation} \mathcal L_\alpha(F) = \mathcal L(F) + \alpha \mathcal R (F) \end{equation}
with $\mathcal L$ stable, $\alpha>0$ and  $\mathcal R$ such that  $\partial_{F_0} \mathcal R>0$  for all 0-subflows.
Let $\alpha_n \rightarrow 0^+$ and let $(F_n)_{n\in \NN}$ be a sequence of  $\mathcal L_{\alpha_n}$-minimizing $R$-edgeflows.

\begin{table*}[!t]
\caption{Translation from directed graphs to measurable spaces \label{table:translation}}
\centering
{
\renewcommand{\arraystretch}{1}
\scalebox{0.8}{
\begin{tabular}[t]{ l|l l l  }
\toprule
& \citet{bengio2021gflownet} &  \citet{lahlou2023theory}&Measurable (ours)\\
 \midrule
State space $S$& DAG &  Topological space, Borel $\sigma$-algebra & Measurable space\\

  Edgeflow & Function on edges  $F(s\rightarrow s')$    & Implicit $\Fout\otimes \ForwardPolicy$&  Measure $F$ on $S\times S$ \\

  In/Outgoingflow     &  Functions on  states  $F_{\text{in}/\text{out}}(s)$  &Measures $\Fin,\Fout$ on $S$&  Idem \\

 Trajectory Flow & Function on trajectories & Measure on trajectories & Idem\\
Reward   &  Function on  states $R(s)$  & Measure $R$ on $S$&  Idem \\

  Forward policy &  $\displaystyle\mathbb P(s_{t+1}| s_{t})=\frac{F(s_t\rightarrow s_{t+1})}{\sum_{s_t\rightarrow s} F(s_t\rightarrow s)}$  &Markov kernel $\ForwardPolicy:S\rightarrow~S$ & $\displaystyle\ForwardPolicy(x,A) = \frac{\mathrm{d}  F(\cdot\rightarrow A)}{\mathrm{d} F(\cdot \rightarrow S)}$\\
 Backward policy &  $\displaystyle\mathbb P(s_{t}| s_{t+1})=\frac{F(s_t\rightarrow s_{t+1})}{\sum_{s\rightarrow s_{t+1}} F(s\rightarrow s_{t+1})}$  &  Markov kernel $\BackwardPolicy:S\rightarrow~S$& $\displaystyle\BackwardPolicy(x,A) = \frac{\mathrm{d} F(A\rightarrow \cdot)}{\mathrm{d} F(S \rightarrow \cdot)}$  \\

 Transitions & Edges &Dominating kernel $\ForwardPolicy\ll\kappa$ & Dominating measure $F\ll \mu$ \\

 Cycles?& Acyclic & $s_f$ finitely absorbing hence acyclic& 0-flows\\

  Reachable Reward& Connectivity & Ergodicity & Dominating measure $R\ll \nu_{train}$\\

    Paths length & Finite graph: $\tau\leq \#S $& $s_f$ finitely absorbing:  $\tau\leq \tau_{max}$& Finite measure: $\displaystyle\mathbb E(\tau)\leq \frac{\Fout(S^*)}{R(S^*)}$ \\
 \bottomrule
\end{tabular}
}
}
\end{table*}
\renewcommand{\arraystretch}{1}

Assume the sequence $(F_n)_{n\in \NN}$ converges to a flow, then it converges toward an acyclic $R$-flow.
\end{theo}
Theorem~\ref{theo:cv} suggests that regularizations controlling the size of the total edgeflow, such as the norm of the edgeflow matrix on graphs,  might help kill cycles thus shortening the expected sampling time. Theorem \ref{theo:sampling} at the end of the following section also goes in this direction.

\subsection{Beyond Graphs}
\label{sec:theoretic}
In order to generalize GFlowNets from the graph intuition to a general measurable state space setting, instead of considering the outgoing flow for a single point or the edgeflow  of a single edge, we consider flows on domains: $F(A\rightarrow B)$   from a domain $A\subset S$ and toward a domain $B\subset S$. It is readily interpreted in the case of graphs as $\sum_{A\ni s \rightarrow s' \in B}F(s\rightarrow s')$, In/outgoing flows are  $F(S\rightarrow \cdot)$ and $F(\cdot \rightarrow S)$, respectively.

In the same way, the reward is considered as a function returning the total reward available some domain $A\subset S^*$. On graphs this means $R(A)=\sum_{s\in A}R(s)$.
On finite graphs, as functions of domains, $R$ and $F_{\text{in/out}}$ are clearly additive e.g. $R(A\cup B) = R(A)+R(B)$ if $A,B$ are disjoint subsets of $S^*$, on the other hand, the edgeflow $F(\cdot\rightarrow \cdot)$ is additive in both variables. A natural extension of the notion flows  is then via measures: an edgeflow is a bimeasure on  $S$ i.e. a measure on $S\times S$ while $F_{\mathrm{in/out}}$ and $R$ are measures on $S$. For technical reasons, we shall restrict to  \emph{finite non-negative measures}: all measure considered $F,\Fout,\Fin, R,...$ are assumed finite. The usual generalization of forward policies of Markov chains in the measurable context is then via Markov kernels. Appendix \ref{sec:reminder_measure} is dedicated to reminders on those objects.

From an edgeflow $F$ as a measure on $S\times S$,  we may  define the outgoing/ingoing flow from/to a given domain $A$ by $\Fout(A):=F(A\rightarrow S)$ and $\Fin(A):=F(S\rightarrow A)$.
Both the {\it reward constraint} for a reward measure $R$ on $S$ and the {\it flow-matching constraint} translate readily to equalities between measures:  $F(\cdot \rightarrow \Final) = R$ and $\mathbf 1_{S^*}\Fin = \mathbf{1}_{S^*}\Fout$.
An edgeflow is an  $R$-edgeflow  if it satisfies the reward constraint for $R$ and an $R$-flow if it satisfies both the reward constraint for $R$ and the flow-matching constraint. In particular, a $0$-flow is a flow satisfying the reward constraint for $R=0$.

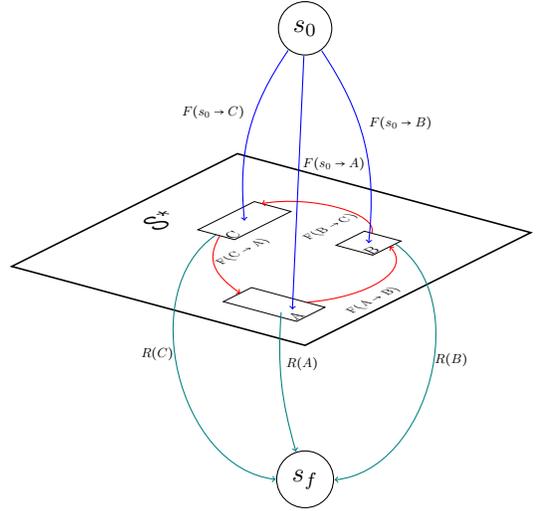
\begin{figure}[h]
    \centering
\scalebox{0.5}{
\begin{tikzpicture}[scale=1.5,every node/.style={minimum size=1cm},on grid]

    \begin{scope}[
        yshift=-160,every node/.append style={
        yslant=0.5,xslant=-1.3},yslant=0.5,xslant=-1.3,
        every edge/.style={draw=blue,thick}
                  ]
        \draw[black,very thick] (0,0) rectangle (4,4);

%

\node at (2,3.5) [scale=2] {$\mathcal{S}^*$};

\draw[black] (0.5,0.5) rectangle (1,1.5);
\node at (0.6,0.6) [scale=1] {$A$};

\draw[black] (2.5,1) rectangle (3,1.5);
\node at (2.6,1.1) [scale=1] {$B$};

\draw[black] (2,2.5) rectangle (3,3);
\node at (2.1,2.6) [scale=1] {$C$};

\coordinate (p1) at (0.75,0.75);
\coordinate (p2) at (2.75,1.25);
\coordinate (p3) at (2.5,2.75);
\coordinate (p4) at (2.8,1);
\coordinate (p5) at (2.9,1);
\coordinate (p6) at (2,2.75);
\path [->] (1,0.75) edge[bend right=50,color=red] node[below left ,scale=0.8,color=black] {$F(A\rightarrow B)$}
                  (p4);
\path [->] (3,1.4) edge[bend right=60,color=red] node[below left ,scale=0.8,color=black] {$F(B\rightarrow C)$}
                  (3,2.9);
\path [->] (2,2.7) edge[bend right=20,color=red] node[right ,scale=0.8,color=black] {$F(C\rightarrow A)$}
                  (0.8,1.5);

    \end{scope}
    \begin{scope}[every node/.style={circle,thick,draw}]
    \node[scale=2] (s0) at (0,0) {$s_0$};
    \node[scale=2] (sf) at (0,-8) {$s_f$};
    \end{scope}
    \begin{scope}[every edge/.style={draw=blue,thick}]

  \path [->] ($(p1) + (-0.2,-0.05)$) edge[bend right=10,color=teal] node[above right,scale=1,color=black]{$R(A)$}
         (sf);
    \path [->] ($(p5) + (0,0)$) edge[bend left=70,color=teal] node[above right,scale=1,color=black]{$R(B)$}
         (sf);

    \path [->] ($(p6) + (0,0)$) edge[bend right=70,color=teal] node[above left,scale=1,color=black]{$R(C)$}
         (sf);

         \path [->] (s0) edge node[above right,scale=1]{$F(s_0\rightarrow A)$}
         (p1);

\path [->] (s0) edge[bend left=20] node[above right,scale=1]{$F(s_0\rightarrow B)$}
         (p2);
\path [->] (s0) edge[bend right=20] node[above left,scale=1]{$F(s_0\rightarrow C)$}
         (p3);

\end{scope}

\end{tikzpicture}
}
\caption{\label{fig:flow}  A depiction of an $R$-edgeflow on a measurable space. The initial flow $F(\Init \rightarrow \cdot)$ is in blue and the Reward $R$ is in green. A possible 0-flow is highlighted in red.}
\end{figure}

In order to properly understand the generalization of edges (i.e. transition restrictions) in the measurable setting,  allow us to remind the reader of the following characterization of the domination relation between measures:
$$\forall \nu,\mu \text{ measures on }S, \quad \nu \ll \mu \Leftrightarrow \exists \varphi ,~ \nu = \varphi \mu.$$
Say $S$ is given by a graph, taking $\mu:=  \sum_{s\rightarrow s'}  \delta_{s\rightarrow s'}$, the relation $F\ll \mu $ is equivalent to the existence of some function $\varphi : S\times S \rightarrow \RR$ such that $F = \sum_{s\rightarrow s'}\varphi(s\rightarrow s') \delta_{s\rightarrow s'}$ in which case the function $\varphi$ is exactly the edgeflow, i.e.
$$F =  \sum_{s\rightarrow s'} F(s\rightarrow s') \delta_{s\rightarrow s'}.$$
We thus generalize edges constraining transitions to the data of a measure $\mu$ dominating the edgeflow. 
The source/sink property of $\Init/\Final$ is enforced by assuming $\mu(\Final \rightarrow S^*)= \mu(S\rightarrow \Init) = 0$ and $\mu(\Final\rightarrow \Final)=1$.
The notion of cycle becomes shaky when dealing with measures in general. However, the notion of 0-flow given in the preceding section extends readily hence the definition of stability is identical.

A technical difficulty appears when linking edgeflows to their induced forward (and backward) policies.
A natural way is via  Radon-Nikodym differentiation
and tensor product of measures, see Appendix \ref{sec:forward_policy} for more details. For now, we simply state that any edgeflow $F$ induces a forward (resp. backward) Markov transition kernel well defined on the `support' of $\Fout$ (resp. $\Fin$) and  conversely, from $(\pi_f,\Fout)$ or $(\pi_b,\Fin)$ one may recover the edgeflow $F$:
$$
\xymatrix{
(\Fin,\pi_b)\ar@<-.35ex>[rr]_{\quad~~~~~\otimes}&&F\ar@<-.35ex>[ll]_{\quad~~~~~(\Fin,\frac{d}{d\Fin})}\ar@<.35ex>[rr]^{(\Fout,\frac{d}{d\Fout})\quad~~~~~}
&&(\Fout,\pi_f)\ar@<.35ex>[ll]^{\otimes\quad\quad}
}.
$$


We may now recover the fundamental sampling result of \citet{bengio2021flow}.
 Let $F$ be an edgeflow and let $\ForwardPolicy$ be its induced forward policy. Consider $(s_t)_{t\geq 0}$ the Markov chain of transition kernel $\ForwardPolicy$ and starting at $\Init$, the sampling time of the edgeflow $F$ is the last position before reaching the sink:
\begin{equation}\tau = \max \{t ~|~  s_ t \neq \Final  \}.\end{equation}

 \begin{theo}\label{theo:sampling} Assuming $R\neq 0$, the sampling time  $\tau$ of a $R$-flow is  almost surely finite and
 the sampling distribution is proportional to $R$. More precisely:
\begin{align}
    \displaystyle \mathbb E(\tau) \leq \frac{\Fout(S^*)}{R(S^*)}, ~\text{and}~
 s_{\tau} \sim  \frac{1}{R(S^*)}R.
\end{align}
 \end{theo}

Finally, \citet{bengio2021gflownet} introduced a notion of trajectory flow that is readily translated in our context as a finite non-negative measure on the space of paths from $\Init$ to $\Final$. We denote by $F^\otimes $ the trajectory flow induced by an edge flow $F$ i.e. $\Fout(\Init)$ times the probability distribution of paths obtained by applying the kernel $\pi_f$ repeatedly starting at $\Init$ until one reaches $\Final$.   Figure \ref{table:translation} summarizes the translation from graphs to measurable state spaces.

\subsubsection{Acyclic 0-Flows}
\label{sec:0flows}
Intuitively, cycles tend to induce exploding sampling time i.e. $\mathbb E(\tau)\rightarrow +\infty$ while training.
While cycles introduce this issue in graphs, 
in the continuous setting, they are not the only factor causing this pathological behavior as the following example shows.

 Take $S^*=\{z\in \mathbb C \mid |z|=1\}$ the unit circle, we may consider a Markov chain of forward policy $\pi_f$ on $S^*$ that is perpetually wandering but never cyclic. One may even take $\pi_f$ deterministic taking rotations such as $\pi_f(z)=\exp(2i\pi\theta)z $ with $\theta$ irrationnal. 
It is tempting to consider that cycles are negligible in the continuous case but the continuous case comes with many ergodic transformations such as irrational rotations as above, each inducing many `\emph{almost-cycles}'. 

The above is an instance of acyclic 0-flow. Controlling 0-flows is thus at the very least a theoretical necessity to control that perpetually wandering trajectories and bounding the expected sampling time.

\section{Stable Non-Acyclic Losses}\label{sec:manage_cycles}

\subsection{Training Losses as Generalized Divergences}

Since the flow-matching property is an equality between measures, it is natural to build losses based on divergences.
Define 
\begin{equation}
\label{div-loss}
    \Div_{g,\nu}(\alpha,\beta) := \int_{x\in \mathcal X} g\left(\frac{d \alpha}{d\beta}(x)\right)d\nu(x),
\end{equation}
where $\alpha,\beta,\nu$ are finite measures on some measurable space $\mathcal X$ and $g$ is some non-negative function. Define $\nu_{\mathrm{state}}$, $\nu_{\mathrm{edge}}$, $\nu_{\mathrm{path}}$ as  training distributions on $S^*$, $S\times S$ and  $\mathrm{Path}(\Init,\Final) :=\{\Init\rightarrow s_1\rightarrow \cdots \rightarrow s_\tau\rightarrow \Final,~~\tau\geq1\}$ respectively. Then, we can write generalizations of FM, DB, and TB losses \eqref{FM-loss}-\eqref{TB-loss} via divergence forms:
\begin{align}
    \LossFM(F) &= \Div_{g,\nu_{\mathrm{state}}} \left(\Fin,\Fout\right), \\
    \LossDB(F_f,F_b)  &= \Div_{g,\nu_{\mathrm{edge}}}(
   F_{f},
 F_{b}), \\
  \LossTB(F_f,F_b) &= \Div_{g,\nu_{\mathrm{path}}} (F^{\otimes}_f, F_b^{\otimes})
\end{align}
where $F_f^\otimes$ is the trajectory flow, non-negative finite measure on the space of paths $\mathrm{Path}(\Init,\Final)$, induced by some edgeflow $F_f$. Beware that the trajectory flow $F_b^\otimes$ is induced by a backward kernel. Direct computations on graphs show the choice $g=\log^2$ indeed yields the losses $\LossFM,\LossDB$ and $\LossTB$ given in \eqref{FM-loss}-\eqref{TB-loss}.

In practice, $F_f$ and $F_b$ are built from  $F_f : = \Fout \otimes \pi_f$ and $F_b := \Fout\otimes \pi_b$
and with the convention that $\pi_b(\Final,\cdot) \propto R$ to enforce the reward constraint. In a similar manner, $F_f^\otimes $ and $F_b^{\otimes}$ are built from $ F_f^\otimes  = \Finit \delta_{\Init} \pi_f^{\otimes \tau}$  and $ F_b^\otimes  = R(S^*) \delta_{\Final} \pi_b^{\otimes \tau}$,  where we defined for any kernel $\pi$ the kernel such that $\pi^{\otimes \tau(x)}$ is the distribution of paths starting at $x$ and ending at the first encounter of $\Init$ or $\Final$.  Hence, $\LossFM,\LossDB$ and $\LossTB$ are special cases of \eqref{div-loss}.


Natural variations of these losses are given by $f$-divergences:
\begin{align}
    \mathcal L_{\text{FM},f}(F) &= \Div_{f} \left(\Fin,\Fout\right), \\
    \mathcal L_{\text{DB},f}(F_f,F_b)  &= \Div_{f} (
   F_{f},
 F_{b}), \\
  \mathcal L_{\text{TB},f} (F_f,F_b) &= \Div_{f} (F^{\otimes}_f, F_b^{\otimes})
\end{align}
where $\Div_f$ denotes an $f$-divergence. We allow $f$-divergences to be improper in the sense that we do not assume a priori that $f$  satisfies the usual hypotheses \cite{polyanski}, i.e.  $f$  convex, strictly convex at $1$ with $f(1)=0$.  We want $\Div_f$  to be distance-like on measures not only on probability measures, i.e. for all non-negative finite measures $\alpha,\beta$:  $\Div_f(\alpha,\beta)\geq 0$ and $\Div_f(\alpha,\beta)=0\Rightarrow \alpha=\beta$. Therefore, we assume that 
$\forall x\in \RR_+^*, f(x)\geq 0$ with equality if and only if $x=1$.  In particular,  this hypothesis excludes Kullback-Leibler cross-entropy.

\begin{theo}[Instability of divergence-based losses]\label{theo:weak_diver}
If $\Div_f$ is a proper divergence, then  $\mathcal L_{\text{FM},f}, \mathcal L_{\text{DB},f}$  are \textbf{unstable} except if $\Div_f$ is the total variation\footnote{see Appendix for a more precise statement.}.

The losses $\mathcal L_{\text{FM},g,\nu},\mathcal L_{\text{DB},g,\nu}$ and $\mathcal L_{\text{TB},g,\nu}$ are unstable.

\end{theo}

\subsection{A  Family of Stable Losses}

The problem of the divergence-like losses is their form  of scale invariance: $t\mapsto \frac{d (F_1+tF_0)}{ d(F_2+tF_0)}$ is converging to $1$ as  $t\geq 0$ grows. This suggests replacing this scale invariance with a $0$-flow invariance: 
$$ \Delta_{f,g,\nu}(\alpha,\beta) := \int_{x\in A} f\left(\alpha(x)-\beta(x)\right)g(\alpha(x),\beta(x))d\nu(x)$$
where $\alpha,\beta$ are measurable functions that will be taken densities of considered measures with respect to a background measure $\lambda$. Then, we can have
\begin{align}
    \mathcal L_{\text{FM}, \Delta,f,g,\nu}(F) &= \Delta_{f,g,\nu_{\mathrm{state}}} \left(\Fin,\Fout\right), \\
    \mathcal L_{\text{DB}, \Delta,f,g,\nu}(F_f,F_b)  &= \Delta_{f,g,\nu_{\mathrm{edge}}}(
   F_{f},
 F_{b}).
\end{align}
In Theorem~\ref{theo:stable_AB}, we prove stable conditions of the above two loss functions. 

\begin{theo}\label{theo:stable_AB} If $f,g,\nu,R$ satisfy
\begin{itemize}
    \item $f\geq 0$ and $g\geq 1$;
    \item $f(x)=0\Leftrightarrow x=0$;
    \item $f$ and $g$ are continuous and piecewise continuously differentiable;
    \item $f$ decreases on $\RR_-$ and increases on $\RR_+$;
    \item $\partial_{(1,1)}g \geq 0$,
\end{itemize}
then for $R$-edgeflows the loss $\mathcal L_{\text{FM},\Delta, f,g, \nu },\mathcal L_{\text{DB},\Delta, f,g, \nu }$ are stable.
\end{theo}

According to theoretical results, we can have the following stable loss examples. It is a pity that we cannot find the stable condition of TB, because the related theoretical analysis of TB is very tricky due to the nonlinearity of the operator $F\mapsto F^{\otimes}$ which returns the trajectory flow associated with an edgeflow. 
We also give a stable version of CFlowNet loss~\cite{li2023cflownets}, which is a variant of FM loss.

\begin{example}[Stable FM loss]
For  $\mathcal L_{FM, \Delta}$, a particularly natural choice of is $f(x)=\log(1+\epsilon |x|^\alpha)$ and $g(x,y) = (1+\eta  (x+y))^\beta$; yielding in the instance of graphs losses:
\begin{align}
    \mathcal L &:=  \mathbb E \sum_{t=1}^\tau \bigg\{ \log\left[1+\varepsilon\left|\Fin(s_t)-\Fout(s_t)\right|^\alpha\right] \nonumber \\ 
    &~~~~~~~~~~~~~~~~~~~~~~~~\times (1+  \eta(\Fin(s_t)+\Fout(s_t)) )^\beta \bigg\} 
\end{align}
for some $\epsilon,\eta,\alpha,\beta>0$. In particular, $(\alpha,\beta,\epsilon,\eta)=(2,1,0.001,1)$ is straightforward as convexity should help convergence for such an under-determined linear problem and dampening too large flow matching error keep the intuitions of the original Bengio loss of not driving too much the training by nodes far from satisfying the flow-matching constraint.

\end{example}

\begin{example}[Stable DB loss]
One may  modify the FM loss above to yield a DB loss:
\begin{align}
    \mathcal L &:=  \mathbb E \sum_{t=1}^\tau \bigg\{ \log\left[1+\varepsilon\left|F^{f}(s_t\rightarrow s_{t+1})-F^{b}(s_t\rightarrow s_{t+1})\right|^\alpha\right] \nonumber \\ 
    &~~~~~~~~~~~~~~~~~~~~~~~~~~~~~~~~~~~~~~~~~~~~~~~\times (1+  \eta\Fout(s_t) )^\beta \bigg\} ,
\end{align}
where $F^{f}(s_t\rightarrow s_{t+1}) = \Fout(s_t)\times \pi_f(s_t\rightarrow s_{t+1}) $ and 
 $F^{b}(s_t\rightarrow s_{t+1}) = \Fout(s_{t+1})\times \pi_b(s_t\rightarrow s_{t+1})$ when forward and backward flows are parameterized by a forward and a backward policy respectively.
\end{example}

\begin{example}[Stable CFlowNet loss] A continuous variation may be written for CFlowNets by replacing $F_{\mathrm{in/out}}$ by their densities $f_{i/o}$ with respect to a background measure:
\begin{align}
    \mathcal L &:=  \mathbb E \sum_{t=1}^\tau \bigg\{ \log\left[1+\varepsilon\left|f_{i}(s_t)-f_o(s_t)\right|^\alpha\right] \nonumber \\ 
    &~~~~~~~~~~~~~~~~~~~~~~~~~~~~~~~~\times (1+  \eta(f_{i}(s_t)+f_{o}(s_t)) )^\beta \bigg\}.
\end{align}
\end{example}

\section{Experiments Results}\label{sec:experiments}

\subsection{Results on Hypergrids}
\begin{figure}[!tb]
       \centering
       {\includegraphics[width=\linewidth]{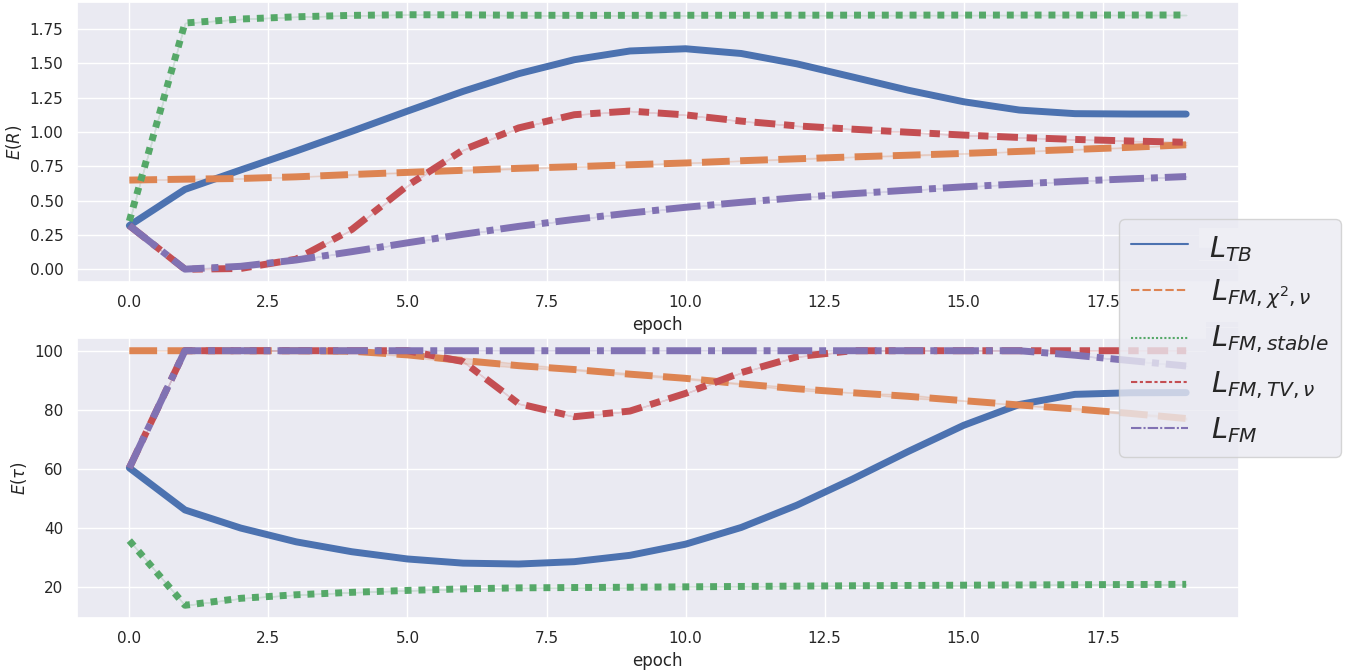}}
    \label{figure:2DW20}
    \caption{\label{fig:HDpaths} GFlowNets were trained with various unstable and stable losses on the 2DW20 grid: $\mathcal L_{FM, stable} = \mathcal L_{FM,\Delta,f,0,\nu}$ with $f(x)=x^2$; $\mathcal L_{FM,\chi^2,\nu}$ and $\mathcal L_{FM, TV,\nu}$ are instances of $\mathcal L_{FM,f,\nu}$ as above with $f(x)=(1-x)^2$ and $f(x)=|1-x|$ respectively.
\textbf{Top}: evolution of averaged reward during training. \textbf{Bottom:} evolution of average lengths of sampled paths.}
\end{figure}

Here $S^*=\lsem 1,W \rsem ^ D $ together with transitions of the form $s \rightarrow s\pm (0,\cdots,0,1,0,\cdots,0)$ in addition to an initial transition
 $\Init\rightarrow a$ for some given $a\in S^*$ and terminal transitions  $\forall s\in S^*, s\rightarrow \Final$. In \citet{bengio2021flow} more restrictive transitions $s \rightarrow s + (0,\cdots,0,1,0,\cdots,0)$  were chosen to enforce acyclicity. GFlowNets for such small graphs were implemented using the adjacency matrix of the underlying graph while the edgeflow along every transition is stored in an array. We compared $\mathcal L_{\text{FM}}$ and $\mathcal L_{\text{TB}}$  to $\mathcal L_{\text{FM},\Delta, f, g}$.
We observed that both losses lead to a converging model but that their paths' behavior differ significantly: the expected path length induced by $\LossFM$ explodes as the paths wander while for $\mathcal L_{\text{FM},\Delta, f, g}$ the paths are relatively straight to reward yielding areas. 
In fact, the edgeflow for non-stable FM losses diverges to infinity reducing the relative reward signal: even if the reward at $s_t$ is high, since the edgeflow toward non-stopping directions are high (due to cycles) then $F(s_t\rightarrow \Final)\ll F(s_t\rightarrow s')$ for $s'\neq  \Final$ and the Markov chain continue to wander, see figures \ref{fig:HGpathsintro} and \ref{fig:HDpaths}. 
The behavior of $\LossTB$ is more nuanced: it induces wandering paths but the expected sampling time does not explode and rather stabilizes at a significantly higher value than for stable losses in some cases. In some others, the TB loss eventually reaches both the same expected reward and expected path lengths as stable losses after a long training time.

\subsection{Results on Cayley Graphs}
Here $S^*=\mathfrak S_p$, the group of permutations of $\lsem 0,p-1\rsem$ for some $p\geq 1$. The edges are given by a set of generators $(\sigma_1,\cdots,\sigma_q)$ so that for all $g\in S^*$ and all $i\in \lsem 1,q\rsem$, we have an edge $g\rightarrow  g\sigma_i$ (more generally one may replace $\mathfrak S_p$ by any discrete group).  $p\geq25$ provides instances of graphs that are intractably big even for current supercomputers but still have small diameters, have tractable FM losses, are easy to implement, and provide a wide variety of structures, symmetries, and cycles.   Many tasks may be reduced to partially ordering a list, we are thus led to construct an element of $\mathfrak S_p$ using generators corresponding to available actions: for instance, a bubble sort only uses transpositions of adjacent objects in the list $L$ to sort, so $p=\mathrm{length}(L)$ and $\sigma_i = (i,i+1)$.  

This experiment differs from the other presented in that the initial policy $\ForwardPolicy(\Init\rightarrow \cdot)$ is fixed as the uniform distribution on $S^*$. Two reasons motivate this choice. Firstly, exploratory behavior is not easy to enforce but we may leverage the highly connected and cyclic nature of Cayley graphs to allow starting from any position. Secondly, it leads to learning a generic strategy to solve a problem from any initial configuration.

\begin{figure}[!tb]
       \centering
       
       {\includegraphics[width=\linewidth]{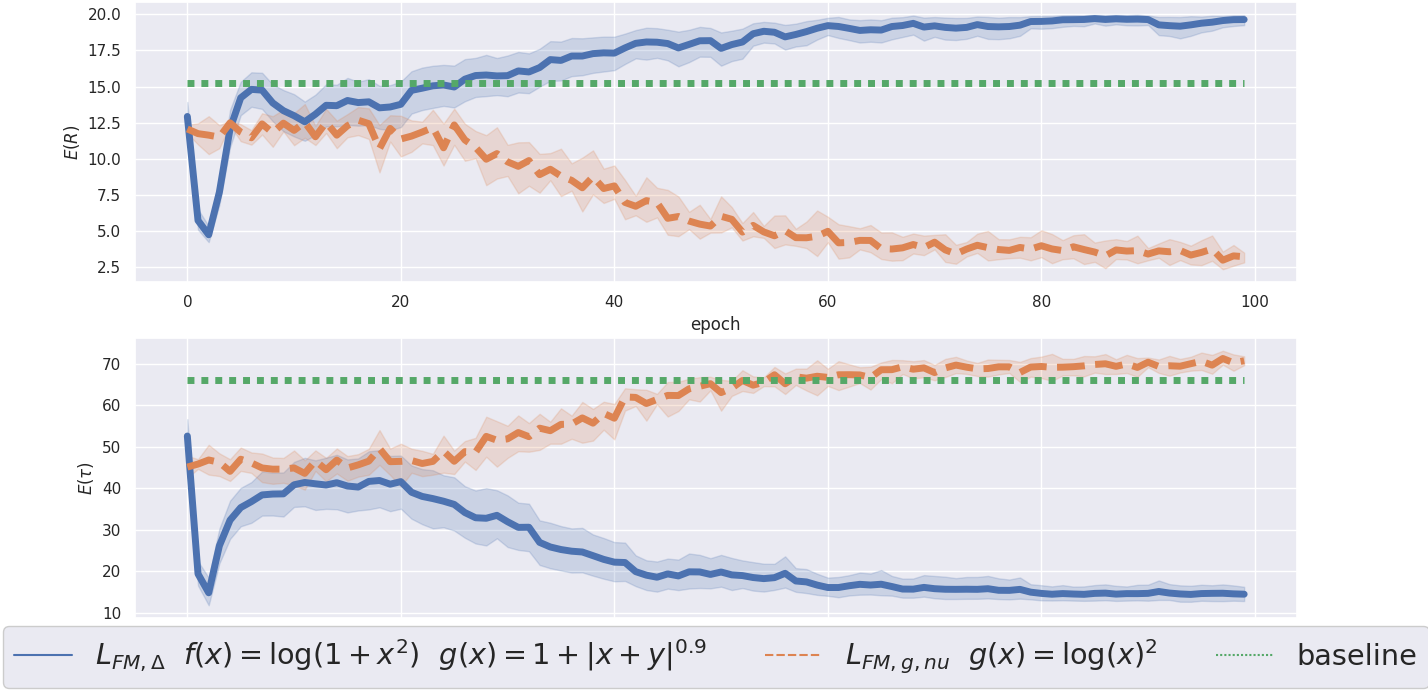}}

\caption{Comparison results of Stable-GFlowNets (ours), GFlowNets, and a Metropolis-Hasting baseline on the Cayley graph of $\mathfrak S_{20}$ generated by a transposition, a $20$-cycle and its inverse. The reward is $R_1$ as above with $k=1$ and $c=20$. Paths are drawn with a cut-off length of 80.   An optimal strategy yields an expected reward of 20 with an expected length of 5. 
\textbf{Top}: evolution of averaged reward under training. \textbf{Bottom:} Average lengths of sampled paths.}
    \label{figure:stable-s20-simple}

\end{figure}

We compared training using $\LossFM$ and $\mathcal L_{\text{FM},\Delta}$ on two families of rewards: $R_1(\sigma):= c\mathbf 1_{\sigma \in S_1} $ and $R_2(\sigma) = d(\rho(\sigma),S_2)$ for some subsets $S_1,S_2\subset S^*$. 
Taking $S_1=\{\sigma ~|~ \sigma(i)=i,i\leq k\}$ leads to emulating a partial sorting algorithm, 
see figure \ref{figure:stable-s20-simple} for $p=20$ and $R_1$ with $c=20$ and $k=1$.

Choices of $R_2$ emulate common heuristics such as variations of the Manhattan distance.  The policy $\ForwardPolicy(\Init\rightarrow \cdot)=\mathcal U(S^*)$ is non-trainable to emulate a random initial instance, thus training the flow to solve a partial ordering problem from any position.
 We observed that, as theoretically predicted, the flow tends to grow uncontrollably when using unstable losses, while using stable losses, the flow stays bounded, thus mitigating the issue of exploding sampling time.


\subsection{Results on Continuous Task}

To show the effectiveness of the proposed Stable-CFlowNets loss, we conduct experiments on Point-Robot-Sparse continuous control tasks with sparse rewards.  The goal of the agent is to navigate two different goals with coordinates $(10,10)$ and $(0,0)$. The agent starts at $(5,5)$ and the maximum episode length is 12. We changed the angle range of the agent's movement from (0, 90°) to (0, 360°), that is, a cycle can be generated in the trajectory. All other experimental settings and algorithm hyperparameters in Roint-Robot-Sparse are the same as in \citet{li2023cflownets}. We refer to \citet{li2023cflownets} for more details. As for Stable-CFlowNets, we only change the loss of CFlowNets to that in Example~3.

\begin{figure}[!tb]
       \centering
       
       {\includegraphics[width=1.65in]{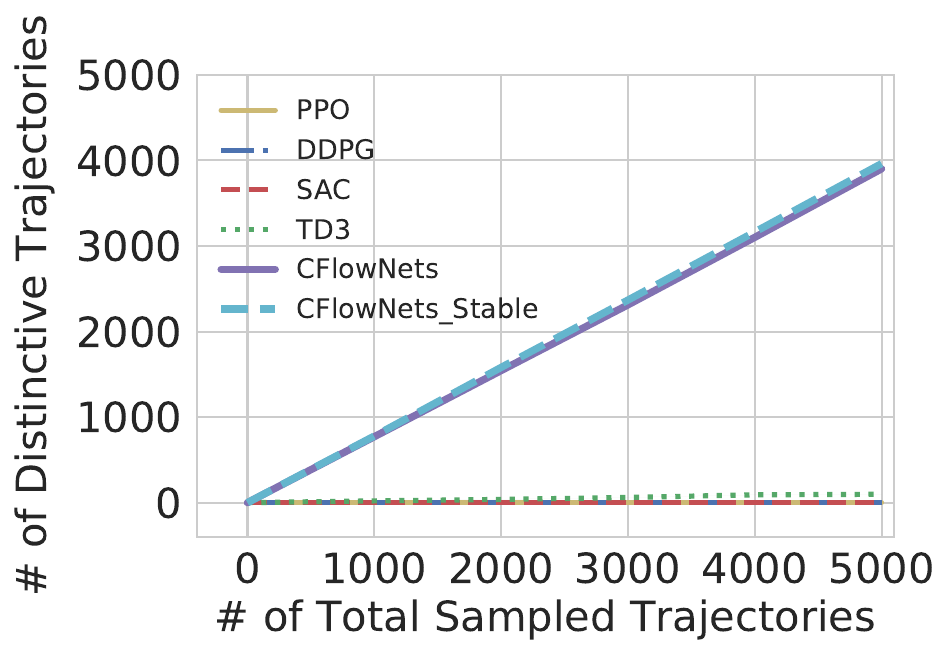}}
       {\includegraphics[width=1.55in]{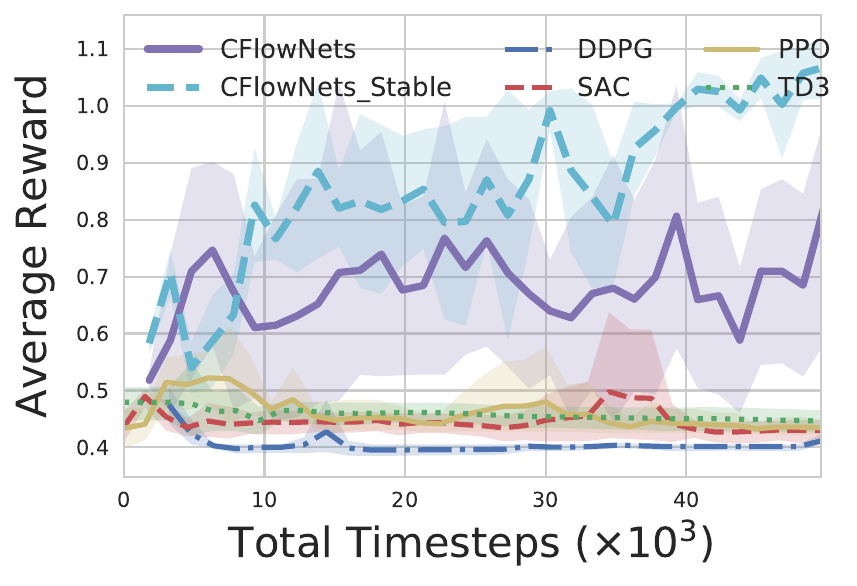}}

	\caption{Comparison results of Stable-CFlowNets (ours), CFlowNets, DDPG, TD3, SAC and PPO on Point-Robot-Sparse. \textbf{Left}: Number of valid-distinctive trajectories generated under 5000 explorations. \textbf{Right:} The average reward of different methods.}
    \label{figure:stable-cfn-reward}
	
\end{figure}

Figure~\ref{figure:stable-cfn-reward} shows the number of valid-distinctive trajectories explored and the reward during the training process. After a certain number of training epochs, 5000 trajectories are collected. We can see that the proposed stable loss does not affect the exploration performance, while the reward has improved a lot, and it has become more stable (the variance range is smaller). It is worth noting that the two completely opposite goals are a difficult environment for RL algorithms, so the reward of the RL algorithms is not high.

\begin{figure}[!tb]
       \centering
       
       {\includegraphics[width=1.55in]{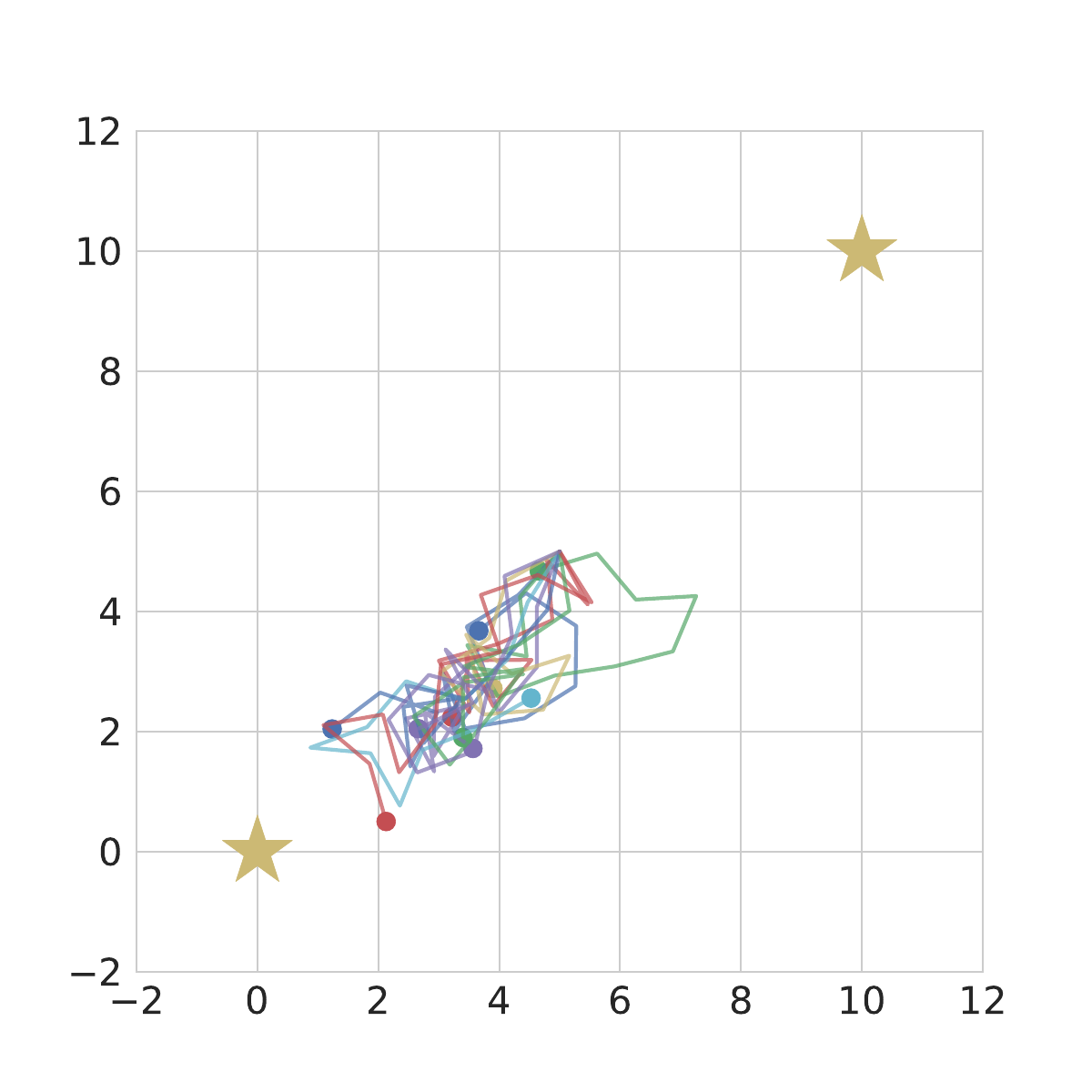}}
       {\includegraphics[width=1.55in]{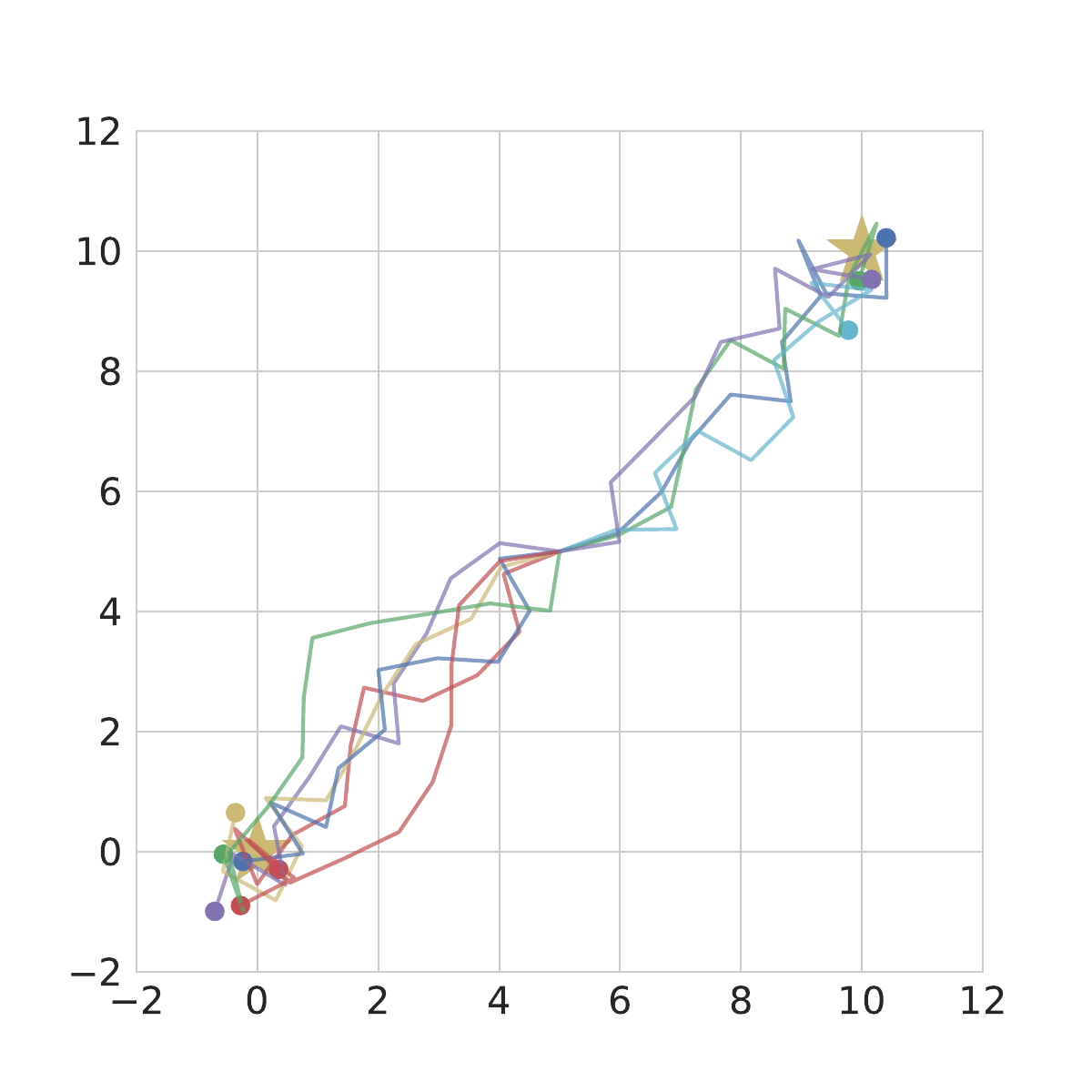}}

	\caption{An example of sampled trajectories on Point-Robot-Sparse task of Stable-CFlowNets (ours). \textbf{Left}: at the beginning of training. \textbf{Right:} at the end of training.}
    \label{figure:stable-cfn-trajectory}
\end{figure}

Figure~\ref{figure:stable-cfn-trajectory} gives an example of sampled trajectories on the Point-Robot-Sparse task of Stable-CFlowNets. To obtain these trajectories, we sample 1000 random actions per node to obtain the corresponding flow outputs (probabilities) and then randomly select one of the top-300 most probable actions to generate trajectories. We can see that there are many cycles in the trajectory at the beginning of the training, but it can still converge to the goal at the end of the training, which shows that our stable loss is very robust to cycles.

\section{Conclusion and Discussion}

We introduced a theoretical framework to generalize GFlowNets concepts and Theorems beyond their initial scope of directed acyclic graphs to measurable spaces. The analysis conducted allowed to link sampling time i.e. the length of sampled paths and the stability of the training via gradient descent. This analysis allowed us to define losses adapted to spaces with possible cycles as well as regularization, helping manage cycles.

\subsection{Related  Works}

In the midst of the growing literature on GFlowNets \cite{bengio2021gflownet,jain2022biological,malkin2022trajectory,madan2022learning,zhang2023robust,jain2022multi,pan2023stochastic,deleu2023generative} three, in particular, are closely related to ours.
\citet{bengio2021gflownet} attempt at laying the foundation of the theory of GFlowNets, they discuss many openings for future applications or explorations of the method.
The work of \citet{lahlou2023theory} using a similar approach to generalize GFlowNets. Finally,  \citet{li2023cflownets} made a first attempt at training GFlowNets for continuous state space. However, none of these works attack cyclic space limitations, in particular, our stability property is new. Furthermore, our framework is somewhat less involved than that of \citet{lahlou2023theory} in that most of the fundamental work deals with general finite non-negative measures. Extra hypotheses enforcing acyclicity are not used nor even specified. 
Furthermore, we explore the structure of the space of flows on graphs, and in general, this is new and a step toward a better understanding of good practices for the training of GFlowNets.

Considering GFlowNets in such a general setting leads to a direct comparison to MCMC \cite{brooks2011handbook} and reinforcement learning \cite{sutton2018reinforcement}.
On the one hand, from a sampling viewpoint, GFlowNets and MCMC are interchangeable: both work in the setting of an unnormalized distribution on a space with the goal of sampling the space proportionally to this target distribution. More precisely, the trainable nature of GFlowNets has to be compared to adaptive MCMC \cite{andrieu2008tutorial}. Two key differences allow us to hope for the replacement of MCMC methods by GFlowNets at least in some situations: GFlowNets benefit from finite mixing-time by opposition of the infinite mixing-time of MCMC, the latter have difficulties moving from modes to modes even after having discovered them while the former can learn sequences of actions to perform such moves.
Mode transition is a long-standing problem impairing MCMC efficiency \cite{andricioaei2001smart,pompe2020framework,samsonov2021local}; one hopes that GFlowNets can provide global moves to help overcome this issue.
On the other hand, GFlowNets on graphs may be thought of as a distributional Q-method \cite{bellemare2017distributional}, the outgoing flow playing the role of the expected reward.

\subsection{Limitations}

First, the hypotheses we used on our instability/stability results are not optimal, and they are technical to allow the use of simple arguments. Furthermore, trajectory balance loss was only partially studied: while FM and DB losses are proved to be unconditionally unstable, TB loss is only unstable in some cases and our analysis and experiments suggest it may be stable for small flows.

Second, our theoretical analysis of Trajectory Balance losses is limited. No stable losses variant of this loss is proposed and our theoretical instability result is rather limited. It suggests that it is possible to find constraints, regularizations, or better parameterizations of flows stabilizing trajectory balance losses. 

Third, the experiments conducted are rather limited compared to the theoretical range of the work. We did not compare stable versions of detailed balance losses in the continuous setting. More thorough experiments testing the influence of non-cyclic $0$-flows in the continuous setting are needed.

Fourth, untrained GFlowNets may have difficulty exploring the space efficiently to learn the target distribution. For adaptive MCMC methods, criteria are known to ensure ergodicity \cite{andrieu2008tutorial}. However, the ergodicity of exploration with GFlowNets is completely open and is expected to suffer from caveats similar to that of adaptive MCMC. Our naive approach was to add a constant edgeflow when sampling paths or add a small background reward.

Fifth, on Cayley graphs, we were not able to train GFlowNets so that the initial flow converges quickly toward the theoretical value. Our simple implementation using an initial flow as a trainable parameter of the model proved to be insufficient.

Finally, our experiments on Cayley graphs used the simplest feedforward neural architecture. More sophisticated architectures taking into account the local graph structure and/or learning more global properties of the graph would be desirable. Graph transformer \cite{dwivedi2020generalization} would be a natural next step, a step already taken on DAG \cite{zhang2023robust} using topoformers \cite{gagrani2022neural}.

\bibliographystyle{icml2023}
\bibliography{note}

\appendix
\section{Extra Theoretical Results}

\subsection{Measures and Markov Kernels}\label{sec:reminder_measure}
Recall that a measurable space \cite{bogachev2007measure} is a set $S$ together with a $\sigma$-algebra of $\mathcal S$ of so-called measurable subsets. For the sake of simplicity, $\mathcal S$ will be assumed implicitly.\footnote{A $\sigma$-algebra $\mathcal S$ of $S$ is a subset of the power set $\mathcal P(S)$ containing $\emptyset$, stable by countable unions and complement.
 For our purpose, either $S$ is countable with $\mathcal S = 2^S$ its power set or $S$ is a Polish space (such as a convex subdomain of $\RR^p$ for some $p\in \NN$) together with $\mathcal S=\mathcal B(S)$ the $\sigma$-algebra of its Borel subsets.}
 A measure on $S$ is a $\sigma$-additive  map
 $\mu:\mathcal S\rightarrow \RR$ \footnote{Meaning $\mu(\bigcup_{i\in I}A_i) = \sum_{i\in I}A_i$ if $(A_i)_{i\in I}$ is a countable family of disjoint measurable subsets.}. \textbf{All measures considered are finite.} We denote by $\delta_s$ the dirac mass at $s$ e.g. the measure such that $\delta_s(A)=1$ if $s\in A$ and other otherwise.

 A Markov kernel from $S$ to $S$ is a measurable \footnote{The space of measures is endowed with the topology of narrow convergence, which is metrizable whenever $S$ is Polish e.g. metrizable with a complete metric; under this assumption, measures form a Polish space \cite{ambrosio2006gradient}. All $S$ of interest for the present work is Polish. We can thus endow $\mathcal M(S)$ with the $\sigma$-algebra of Borel subsets for this topology to obtain a reasonable measurable space.
Furthermore,  all maps in the scope of the present work are measurable.} map $\pi:S\rightarrow \mathcal M(S)$ such that $\pi(x \rightarrow S)=1$ for all $x\in S$ where $\mathcal M$ denote the set of finite non-negative measures. Markov kernels are the natural setting for Markov chains \cite{douc2018markov}.
  For notation convenience, we write $\pi(x \rightarrow A)$ the measure of $A$ for the measure $\pi(x)$.
  A Markov kernel can be seen as a random map \cite{kifer2012ergodic}.

  Such a Markov kernel $\pi$ acts on  measures $\mu \pi (A) = \int_{x\in S} \pi(x \rightarrow A) d\mu(x)$ the special case where $\mu$ is a probability measure can be interpreted as  $\mu \pi(A)=\mathbb P_{x\sim \mu}(\pi(x) \in A)$ viewing $\pi$ as a random map. The tensor product of a measure $\mu$ by a kernel $\pi$ is the measure on $S\times S$ given by $\mu\otimes \pi (A\times B) := \int_{x\in A} \pi(x \rightarrow B)d \mu (x)$ for all measurable $A,B\subset S$.

  In practice, a measure $\mu$ is constructed as a density with respect to a reference dominating measure $\lambda$ i.e. $\mu = \varphi \lambda$ for some (measurable) function $\varphi$.
  Such domination is denoted $\mu \ll \lambda$ and by the Radon-Nikodym Theorem is equivalent for finite measures to the following property:
  $$\forall B\in \mathcal S,\quad \lambda(B)=0\Rightarrow \mu(B)=0.$$
In the whole section,  $S$ is a measurable space with two special points $\Init,\Final$, we denote by $S^*:=S\setminus \{\Init, \Final\}$. We also give ourselves a transition restriction $\mu$ i.e. a measure on $S\times S$ such that $\mu(\cdot\rightarrow \Init)=\mu(\Final\rightarrow \cdot)=0$.

\subsection{Edgeflows Versus Weighted Markov Chains}
\label{sec:forward_policy}
One may build GFlowNets in two ways, either as a weighted Markov chain $(\ForwardPolicy,\Fout)$ or as an  edgeflow $F(\cdot\rightarrow \cdot)$.
The following proposition shows one can reconstruct the weighted Markov kernel from edgeflow  and vice versa.
\begin{defi}
Let $({\ForwardPolicy}_i,G_i)_{i\in \{1,2\}}$ be  two weighted Markov kernels. They are equivalent if $G_1=G_2$ and
\begin{equation}
 {\ForwardPolicy}_1(x\rightarrow B)={\ForwardPolicy}_2(x \rightarrow B)\end{equation}
                   $G_1(x)$-almost everywhere for all measurable $B\subset S$.
We denote this equivalence by $({\ForwardPolicy}_1,G_1)\sim ({\ForwardPolicy}_2,G_2)$.
\end{defi}

 \begin{prop}\label{prop:edgeflow_base} Weighted Markov kernels $(\ForwardPolicy,G)$  or measures on $S\times S$ may be used indeferently to define edgeflows. More precisely:

\begin{enumerate}[(a)]
\item For all non-negative finite measure $F$, then $F=\Fout \otimes \ForwardPolicy^F$ where for all $A\subset S$:
\begin{equation}\ForwardPolicy^F(\cdot\rightarrow A) := \frac{d F(\cdot \rightarrow A)}{d F(\cdot \rightarrow S)}.\end{equation}
Furthermore, for two such measures $F_1,F_2$, if $F_1\ll F_2 $ then  $F_{1,\mathrm{out}}\ll F_{2,\mathrm{out}}$ and $$\pi_f^{F_1}(x)\ll \pi_f^{F_2}(x)$$
for $F_2(x)$-almost all $x$.

\item  For any weighted Markov kernel $(\ForwardPolicy,G)$ we have
$$(\ForwardPolicy,G) \sim (\ForwardPolicy^{G\otimes \ForwardPolicy},G\otimes \ForwardPolicy(\cdot\rightarrow S)).$$
Furthermore, for any two weighted Markov kernels $(\pi_f^1,G_1),(\pi_f^2,G_2)$ if
$G_1\ll G_2$ and $\pi_1(x)\ll \pi_2(x)$ for $G_2$-almost all $x$ then $$ G_1\otimes \pi_f^1 \ll G_2 \otimes \pi_f^2.$$
\end{enumerate}
 \end{prop}
 Similar results may be stated linking the backward policy to the edgeflow. 
    Beware that we denote tensor products ``measure $\otimes$ forward policy" and ``measure $\otimes$ backward policy"  the same way, however:
    $$F(A\rightarrow B) = \Fout\otimes \pi_f(A\rightarrow B) = 
    \Fin\otimes \pi_b(B\rightarrow A).$$
    Since, from context, one knows if a policy is ``forward" or ``backward", we allow ourselves this imprecision.

The consequence of this last Proposition are the following (among others).
\begin{itemize}
    \item In the framework of \citet{lahlou2023theory} the transition constraints are enforced via a domination of the forward policies by a reference transition kernel; the proposition shows that, together with a domination of the outgoing flow, this is equivalent to our domination of edgeflows.
    \item Since the transition constraint $\mu$ on $S\times S$ is such that $\mu(s_f\rightarrow \cdot)= \delta_{\Final} $, the edgeflow induced kernel  $\ForwardPolicy^F$ at $\Final$, taking the convention that $\ForwardPolicy^F$ is a Markov Kernel for all $x\in S$ we have that $\ForwardPolicy^F(\Final \rightarrow \Final)=1$.
\end{itemize}

\subsection{Sampler Flow and Expected Sampling Time}\label{sec:sampler_flow}

 When sampling the flow, either during training or during inference, one is mainly interested in the ``particle flow'' e.g. the flow given by infinitely many paths sampled with the forward policy $\ForwardPolicy$ induced by $F$. This merits a proper definition.

\begin{defi}[Sampler Flow]
The sampler flow of an edgeflow $F$ of forward policy $\pi_f$ and such that $\mathbb E(\tau)<+\infty$ is the edgeflow $\overline F=\Foutbar\otimes \ForwardPolicy$ with
\begin{equation}\Foutbar (A) := \Fout(\Init)\times \mathbb E \left(\mathrm{card}\{t\in  \NN~|~ \Final \neq s_t\in A\}\right ).\end{equation}
An edgeflow is exactly sampled if $\overline F=F$.
\end{defi}
\begin{prop}\label{lem:reach_flow} Let $F$ be an edgeflow such that $\mathbb E(\tau)<+\infty$,
then $\overline F$ is a flow.
If furthermore $F$ is a $R$-flow then $\overline F$ is a $R$-flow and $\overline F\leq F$.
\end{prop}
\begin{cor}\label{cor:sampling_time_bound} The total flow bounds the expected sampling time: 
$$ \mathbb E(\tau) = \frac{\overline F(S^*\rightarrow S^*)}{R(S^*)} \leq  \frac{F(S^*\rightarrow S^*)}{R(S^*)}.$$
\end{cor}

\begin{figure}[!h]
    \centering

$$
\xymatrix{
 \Init  \ar@<.35ex>[r]^1  \ar@<-.35ex>@[red][r]_{\color{red} 1}  &A\ar@<-.35ex>@[red][r]_{\color{red} 1}  \ar@<0.35ex>[r]^2 &\Final&B \ar[l]_{1} \\
}
$$
$$
\xymatrix{
 \Init  \ar@<.35ex>[r]^1  \ar@<-.35ex>@[red][r]_{\color{red} 1}  &A\ar@<-.35ex>@[red][r]_{\color{red} 1}  \ar@<0.35ex>[r]^1 &\Final  &B \ar@/_1pc/[r]^1 &C  \ar@/_1pc/[l]^1
}
$$
    \caption{ edgeflows (black) and associated sampling flows (red)}
    \label{fig:F_Fbar}
\end{figure}
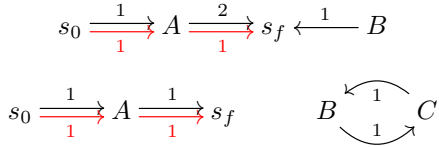

   From Theorem \ref{theo:sampling},  denoting $\overline R:=\overline F(\cdot\rightarrow \Final)$ we have  $s_\tau \sim \frac{1}{\overline R(S^*)}\overline R$
  for any edgeflow such that $\mathbb E(\tau)<+\infty$.
  Figure \ref{fig:F_Fbar} represents the difference between an edgeflow and its induced sampler flow.

 \subsection{Flows of Cyclic Type}
 Although $0$-flows generalize cycles on graphs, it is unclear whether all $0$-flows are sums of cycles. We give a rigorous definition of ``sum of cycles'' in the measurable context together with a counter-example.
 \begin{defi}
For $\gamma = (s_0,\cdots,s_{\ell-1})$ a length $\ell$ cycle in $S^*$ define the 0-flow $\mathbf{1}_{\gamma}=\sum_{i\in \Z/\ell \Z} \delta_{s_i}\otimes \delta_{s_{i+1}}$.
Let $\Gamma(S^*)$ be the set of cycles $(s_0,\cdots,s_{\ell-1})$ of arbitrary length $\ell\in \NN^*$

A (resp. generalized) 0-flow  $F$,  is of cyclic type  if it admits decomposition

$$ F = \int_{\gamma}\mathbf{1}_\gamma \mathrm {d} \mu(\gamma)$$
for some $\mu$ (resp. signed) finite measure on $\Gamma(S^*)$.
\end{defi}
\begin{defi} A edgeflow is acyclic if it admits no non-trivial 0-subflow of cyclic type.

\end{defi}
\begin{prop}\label{prop:decomposition} A edgeflow $F$ has a maximal  0-subflow hence it can be written $$F = F_0+ F_{min}$$
where $F_0$ is a 0-subflow and $F_{min}$ is a minimal edgeflow.
\end{prop}

We shall show in Section \ref{sec:graphs} that on countable state spaces, all 0-flows are of cyclic type. This allows us to prove the following.
\begin{lemma} \label{lem:acyclic_close} If $S$ is at most countably infinite, then the subsets of acyclic edgeflows  and minimal edgeflows are both closed for narrow convergence.
\end{lemma}

For a edgeflow, minimal implies acyclic. For a flow, minimal also implies exactly sampled. If $S$ is uncountable, then a flow may be acyclic and exactly sampled but not be minimal.

\begin{ex} The flow depicted on the left contains two distinct maximal 0-subflows (red, blue) depicted on the right together with an associated (to red) minimal subflow (green):
$$
\xymatrix{
 A \ar[r]^1 &D\ar[d]^1 \ar[r]^1&\Final\\
 \Init \ar[r]^1&B\ar[ul]_1 \ar[r]^1 &C\ar[ul]_1
}\quad \quad
\xymatrix{
 A \ar@[red][r] &D\ar@[red]@<-0.3ex>[d]\ar@[blue]@<0.3ex>[d]\ar@[green][r]&\Final\\
\Init\ar@[green][r] &B\ar@[red][ul]\ar@[green]@<-0.3ex>[r] \ar@[blue]@<0.3ex>[r] &C\ar@[blue]@<0.3ex>[ul]\ar@[green]@<-0.3ex>[ul] &
}
$$

\end{ex}

\subsection{Structure of Flows Space on Graphs} \label{sec:graphs}

To begin with we notice that a graph $G=(S,E)$ can be seen as an at most countably infinite discrete space $S$ together with a domination on flows by the counting measure on $E$ e.g. $F\ll \mu_{E}:=\sum_{(s,s')\in E}\delta_{s}\otimes \delta_{s'}$.
All the results presented in the previous sections apply to this setting, we thus have a sampling theorem, an upper bound on the expected sampling time, a decomposition property, and a choice of flow-matching losses.
We assume that $G$ is connected in the sense that for every state $s$, there exists a path from $\Init$ to $\Final$ through $s$.

Consider a topological\footnote{Here topological means that edges can be crossed forward or backward.} simple loop $\Gamma$ in $G$:  $\gamma= s_0\rightarrow s_1 \leftarrow s_2 \leftarrow s_3 \rightarrow s_4 \leftarrow \cdots \rightarrow s_0$ so that states appear once (except for the first which appears twice); we define
$$\mathbf{1}_\gamma(s\rightarrow s') =  \left\{ \begin{array}{cc} \displaystyle 1 & \text{If } s\rightarrow s' \in \gamma \\  \displaystyle -1 & \text{If } s'\rightarrow s \in \gamma  \\ \displaystyle  0 & \text{otherwise} \end{array} \right.$$
then $\mathbf{1}_\gamma$ is a generalized flow.

 $\mathcal F(G),\mathcal F^+(G),\mathcal F_R(G),\mathcal F^+_R(G) $  denote respectively the space of generalized flows, of flows, of generalized $R$-flows and of $R$-flows. We denote by $H^1(G)$ the closure\footnote{In the context of graphs, the space of generalized flows $\mathcal F(G)$  on $G=(S,E)$ is the space of summable flows e.g. functions  $F: E\rightarrow \RR$ such that $\|F\|_1 := \sum_{e\in E}|F(e)|<+\infty$ and satisfying the Flow Matching property. One recognizes $\mathcal F(G)\subset\ell^1(E)=L^1(E,\mu_{E})$.} in $\mathcal F(G)$ of the vector space spanned by such $\mathrm 1_{\gamma}$  with $\gamma \subset G\setminus \{\Final\}$.

\begin{prop}\label{prop:graph_dec_generalized} Let $G=(S,E)$ be a connected directed graph and let $R$ be a summable target distribution.  Then $\mathcal F_R$ is a non-empty affine subspace of $\ell^1(E)$ :
$$ \mathcal F_R = F + H^ 1(G) $$ where $F$ is any element of $\mathcal F_R$. In particular, $\mathcal F_0 = H^1(G)$.
\end{prop}
 A proof is included in the appendix; however, this proposition follows from Theorem 3.3.1 from \cite{kalpazidou2007cycle}, which provides a stronger result.
\begin{cor} All generalized 0-flows on graphs are of cyclic type.
\end{cor}
%
%
%

The following proposition extends this property to non-negative flows.  To this end we introduce $ \mathcal F^+_{R,acyclic} \subset  \mathcal F^+_{R}$  composed of non-negative acyclic flows and
$H^1_+(G)$ the closure of $\bigoplus_{\gamma} \RR_+ \mathbf{1}_\gamma\subset H^1(G)$   where $\gamma$ run through simple loops of $G$ (instead of merely topological simple loops so that $\mathbf 1_{\gamma}\geq 0$).

\begin{prop}\label{prop:graph_dec} Let $G=(S,E)$ be a connected directed graph and let $R$ non-negative reward function.  Then $\mathcal F_R^+$ is a  non-empty convex domain of $\mathcal F_R$.
Furthermore,
$$ \mathcal F^+_R = \mathcal F^+_{R,acyclic} + H^1_+(G).$$
\end{prop}
\begin{cor} All 0-flows on graphs are of cyclic type.
\end{cor}

\section{Proofs}

\begin{repprop}{prop:edgeflow_base}
Weighted Markov kernels $(\ForwardPolicy,G)$  or measures on $S\times S$ may be used indifferently to define edgeflows. More precisely:

\begin{enumerate}[(a)]
\item For all non-negative finite measure $F$, then $F=\Fout \otimes \ForwardPolicy^F$ where for all $A\subset S$:
\begin{equation}\ForwardPolicy^F(\cdot\rightarrow A) := \frac{d F(\cdot \rightarrow A)}{d F(\cdot \rightarrow S)}.\end{equation}
Furthermore, for two such measures $F_1,F_2$, if $F_1\ll F_2 $ then  $F_{1,\mathrm{out}}\ll F_{2,\mathrm{out}}$ and $$\pi_f^{F_1}(x)\ll \pi_f^{F_2}(x)$$
for $F_2(x)$-almost all $x$.

\item  For any weighted Markov kernel $(\ForwardPolicy,G)$ we have
$$(\ForwardPolicy,G) \sim (\ForwardPolicy^{G\otimes \ForwardPolicy},G\otimes \ForwardPolicy(\cdot\rightarrow S)).$$
Furthermore, for any two weighted Markov kernels $(\pi_f^1,G_1),(\pi_f^2,G_2)$ if
$G_1\ll G_2$ and $\pi_1(x)\ll \pi_2(x)$ for $G_2$-almost all $x$ then $$ G_1\otimes \pi_f^1 \ll G_2 \otimes \pi_f^2.$$
\end{enumerate}
\end{repprop}
 \begin{proof}
 \begin{enumerate}[(a)]
\item Since $F\geq 0$, we have $\forall A\in \mathcal S, F(\cdot \rightarrow A )\leq F(\cdot\rightarrow S)$ so, by Radon-Nykodym theorem, $\ForwardPolicy^F$ is well defined and  $$\forall A\in\mathcal S, \quad F(\cdot \rightarrow A) = \ForwardPolicy^F(\cdot,A)\Fout.$$
For all $A,B\in \mathcal S$,
\begin{eqnarray}
    \Fout \otimes \ForwardPolicy^F (A,B) &:=& \int_{x\in A} \ForwardPolicy^F(x,B) \mathrm d \Fout(x)~~~~~~~ \\
    &=&(\ForwardPolicy^F(\cdot,B) \Fout) (A) \\ &=& F(A \rightarrow B)
\end{eqnarray}
so  $F= \Fout \otimes \ForwardPolicy^F$.

Let $\varphi\in L_1(F_2)$ such that $F_1 = \varphi F_2$ we have for all $A,B\in \mathcal S$:
\begin{flalign}
 &F_1(A\rightarrow B)=&&\nonumber\\
 &\quad \int_{x\in A}\int_{x\in B} \varphi(x,y) d[\pi_f^{F_2}(x)](y) d F_{2,\mathrm{out}}(x)   &&\nonumber
\end{flalign}

so $\forall B\in \mathcal S, F_1(\cdot \rightarrow B) \ll F_{2,\mathrm{out}}$  and  for $F_{2,\mathrm{out}}$-almost all $x$:
$$\frac{d F_1(\cdot\rightarrow B)}{ d F_{2,\mathrm{out}} }(x) =  \int_{y\in B} \varphi(x,y)d[\pi_f^{F_2}(x)](y).$$
Thus $F_{1,\mathrm{out}}=F_1(\cdot\rightarrow S)\ll F_{2,\mathrm{out}}$ and for $F_{1,\mathrm{out}}$-almost all $x$ and for all $B\in \mathcal S$:
$$  \frac{d F_1(\cdot\rightarrow B)}{ d  F_1(\cdot\rightarrow S) }(x) = \frac{ \int_{y\in B} \varphi(x,y)d[\pi_f^{F_2}(x)](y)}{ \int_{y\in S} \varphi(x,y)d[\pi_f^{F_2}(x)](y)}.$$
In particular, for $F_{1,\mathrm{out}}$-almost all $x$, $$ \pi_f^{F_1}(x\rightarrow \cdot) = \Phi_x \pi_f^{F_2}(x,\cdot ) $$
with $\Phi_x(y)\in L^{1}(\pi_f^{F_2}(x,\cdot))$ given by
$$\Phi_x(y) =  \frac{ \varphi(x,y)}{ \int_{y\in S} \varphi(x,y)d[\pi_f^{F_2}(x)](y)}.$$
 \item Define $\Fout := G\otimes \ForwardPolicy(\cdot \rightarrow S)$,
 for all  $A\in \mathcal S$,
\begin{eqnarray*}
 \Fout(A)&:=& G\otimes \ForwardPolicy(A \rightarrow S) \\
 & =& \int_{x\in A} \ForwardPolicy(x,S) d G(x)  \\
& =& \int_{x\in A} d G(x) = G(A).
 \end{eqnarray*}
  Furthermore, for any $B\subset S$ measurable, we have the following equality as measures:
 $$F(\cdot,B)=\Fout \otimes \ForwardPolicy(\cdot,B) = \ForwardPolicy(\cdot,B) \Fout;$$
 therefore the measurable function $x\mapsto \ForwardPolicy(x,B)$ is a derivative of $ F(\cdot,B)$ with respect to $ \Fout=G$.
 By uniquenes of derivative, we have $G(x)$-almost everywhere  ${\ForwardPolicy}_{G\otimes \ForwardPolicy}(x,B)=\ForwardPolicy(x,B)$, hence $(\ForwardPolicy,G)\sim (\ForwardPolicy^{G\otimes \ForwardPolicy}, \Fout)$.

 Given two such weighted Markov kernels,  clearly:
 $$ G_1\otimes \pi_f^1 = \Phi\times  (G_2\otimes \pi_f^2)$$
 with $\Phi(x,y):=\frac{d G_1}{d G_2}(x)\frac{d \pi_f^1(x)}{d \pi_f^2(x)}(y)$.

  \end{enumerate}
 \end{proof}

  \begin{lemma}\label{lem:flow_rec} Let $F$ be a (generalized) R-edgeflow on $S$, then $F$ is  a (generalized) flow
iff $\displaystyle \Fout {\ForwardPolicy} - \Fout = R(S^*)\delta _{\Final}- \Fout(\Init)\delta_{\Init} $
 \end{lemma}
 \begin{proof}
 If $F$ is a flow then by definition $\mathbf{1}_{S^*}(\Fout {\ForwardPolicy} - \Fout) = 0$; since $F$ is a $R$-edgeflow then $\Fout \ForwardPolicy(S\rightarrow \Final)=F(S\rightarrow \Final)=R(S^*)$, $\ForwardPolicy(\Final\rightarrow \cdot) = \delta_{\Final}$  and $\Fout\pi_f (\Init) = F(S\rightarrow  \Init)=0 $  hence  $\displaystyle \Fout {\ForwardPolicy} - \Fout = R(S^*)\delta _{\Final}- \Fout(\Init)\delta_{\Init}$.

 If  $\displaystyle \Fout {\ForwardPolicy} - \Fout = R(S^*)\delta _{\Final}- \Fout(\Init)\delta_{\Init}$ then in particular
 $\mathbf{1}_{S^*}\Fin = \mathbf{1}_{S^*}\Fout$ so $F$ is a flow. Since it's a $R$-edgeflow, $F$ is a $R$-flow.
 \end{proof}

  \begin{prop}\label{prop:flowspace} The space of generalized $R$-flows for a given $R$ is an closed affine subspace of the space of finite measures on $S\times S$ directed by the closed vector space of generalized $0$-flows.
The subset of  $R$-flows is a closed convex domain.
\end{prop}

\begin{proof}
The Flow Matching constraint is clearly linear and the Reward constraint is clearly affine.  The sum of a generalized $R_1$-flow with a generalized $R_2$-flow is a generalized $(R_1+R_2)$-flow hence the direction of the affine subspace of generalized $R$-flows is the linear space of generalized $0$-flows. Narrow convergence of $F_n\rightarrow F$ implies that $F_n(A \rightarrow B)\xrightarrow{n\rightarrow +\infty} F(A \rightarrow B)$ for all $A,B\in \mathcal S$ so the Flow Matching constraints and Reward constraints goes to the limit, hence $\mathcal F_R$ is closed.

Finally, $\mathcal F_R^+$ is the intersection of the closed affine space $\mathcal F_R$ with the closed convex cone of  non-negative measures.
\end{proof}

\begin{reptheorem}{theo:sampling}
Assuming $R\neq 0$, the sampling time  $\tau$ of a $R$-flow is  almost surely finite and
 the sampling distribution is proportional to $R$. More precisely:
\begin{align}
    \displaystyle \mathbb E(\tau) \leq \frac{\Fout(S^*)}{R(S^*)}, ~\text{and}~
 s_{\tau} \sim  \frac{1}{R(S^*)}R.
\end{align}
\end{reptheorem}
The Theorem is a consequence of the three following propositions together with Corollary \ref{cor:sampling_time_bound}.

\begin{prop}\label{prop:sampling_time} The sampling time  $\tau$ of a $R$-flow is  almost surely finite if   $R\neq0$.
\end{prop}
\begin{proof}
Assume $R\neq 0$. Recall that $ F(\Init) = R(S^*)>0$.
By Lemma \ref{lem:flow_rec}, for all $k\in \NN$ we have
$$ \Fout(\Init) \delta_{\Init} \ForwardPolicy^k = (\Fout \ForwardPolicy^k-\Fout \ForwardPolicy^{k+1}) + F(S^*,\Final)\delta_{\Final}$$
and by definition of $\tau$ we have for all  $k\in \NN^*$ $$ ~~ \mathbb P(\tau> k) = \delta_{\Init}{\ForwardPolicy}^k(S^ *) = \frac{ \Fout \ForwardPolicy^k(S^*)-\Fout \ForwardPolicy^{k+1}(S^*)}{F(\Init)}.$$

Note that the sequence  $\left(\Fout \ForwardPolicy^k(S^ *)\right)_{k\in \NN^*}$  is  non-increasing non-negative, hence convergent. It follows that
$\lim_{k\rightarrow +\infty} \left(\Fout \ForwardPolicy^k(S^*)-\Fout \ForwardPolicy^{k+1}(S^*)\right)=0$.
Therefore $\lim_{k\rightarrow +\infty} \mathbb P(\tau\geq k) = 0$. In other words, $\tau$ is almost surely finite.
\end{proof}

\begin{prop}\label{prop:sampling_dist} Assuming $R\neq 0$, the sampling distribution is proportional to $R$. More precisely:
\begin{align}
 s_{\tau} \sim  \frac{1}{R(S^*)}R.
\end{align}
\end{prop}
\begin{proof}
By Lemma  \ref{lem:flow_rec} the sequence $\Fout \ForwardPolicy^k$ is non-negative,  non-increasing on $S^*$ and non-decreasing on $\Final$, thus converges toward a measure $\mu=\nu+\beta \delta_{\Final}$ with $\beta\in \RR_+\cup \{+\infty\}$ and $\mathrm{supp}(\nu) \subset S^*$.

Furthermore, $\mu\ForwardPolicy = \mu$ and $\delta_{\Final}{\ForwardPolicy}=\delta_{\Final}$  so $$ (\nu+\beta \delta_\Final)\ForwardPolicy =  \nu\ForwardPolicy + \beta \delta_{\Final}  =  \nu + \beta \delta_{\Final}  $$
so $\nu\ForwardPolicy = \nu $ and since $\ForwardPolicy$ is Markovian we deduce that $\nu \otimes \ForwardPolicy(S^*,\Final)=0$.

Then, for every  $A$ measurable subset of $S^*$,
\begin{eqnarray*}
&&\mathbb P(M_{\tau} \in A) \\
&=& \left (  \sum_{k\in \NN}  \delta_{\Init} \ForwardPolicy^k\right) \otimes \ForwardPolicy(A\times\{\Final\}) \\
&=& \left (  \sum_{k\in \NN} \frac{1}{R(S^*)} (\Fout \ForwardPolicy^k-\Fout \ForwardPolicy^{k+1}) + \delta_{\Final}\right)\\
&&~~\otimes \ForwardPolicy(A\times\{\Final\}) \\
&=& \frac{1}{R(S^*)}  \left (\Fout-\nu\right) \otimes \ForwardPolicy(A\times\{\Final\}) \\
&=&  \frac{1}{R(S^*)} F(A\times\{\Final\}) - \frac{1}{R(S^*)} \underbrace{\nu\otimes \ForwardPolicy(A\times\{\Final\})}_{=0} \\
&=& \frac{R(A)}{R(S^*)}.
\end{eqnarray*}
\end{proof}

\begin{repprop}{lem:reach_flow} Let $F$ be an edgeflow,
then $\overline F$ is a flow if and only if  $\mathbb E(\tau)<+\infty$.
If $F$ is a $R$-flow then $\overline F$ is a $R$-flow and $\overline F\leq F$ and 
$$\mathbf{1}_{S^*}\lim_{k\rightarrow+\infty} \overline \Fout \ForwardPolicy^{k}=0.$$
\end{repprop}

\begin{proof}
If the expected sampling time of $F$ is not finite, then $\overline F$ is certainly not a finite measure thus not an edgeflow, thus not a flow. Assume that the expected sampling time of $F$ is finite, it follows that $\overline F$ is a finite non-negative measure on $S\times S$ thus an edgeflow.
Define $\overline R = \overline F(\cdot\rightarrow \Final)$.
Write $$\Foutbar = \Fout(\Init) \mathbf 1_{S\setminus\{\Final\}}\sum_{t=0}^{+\infty} \delta_{\Init}\ForwardPolicy^t.$$
We have  $\Foutbar (\Final) =0$, $ M_0 \sim \delta_{\Init} $ and
$$\Foutbar -\Foutbar {\ForwardPolicy} = \Fout(\Init) \delta_{\Init}-\overline F(S \rightarrow \Final)\delta_{\Final}$$
so $\overline F$ is a $\overline R$-flow by Lemma \ref{lem:flow_rec}.

Assume that $F$ is an $R$-flow.  Then,
\begin{eqnarray*}
  \mathbf 1_{S^*}\Foutbar &=&  \sum_{k=0}^{+\infty} \mathbf 1_{S^*} \Fout(\Init)\delta_{\Init}\ForwardPolicy^k   \\
  &=&  \mathbf 1_{S^*}\sum_{k=0}^{+\infty}  (\Fout \ForwardPolicy^k-\Fout \ForwardPolicy^{k+1}) \\
   &=&  \mathbf 1_{S^*} \left(\Fout - \lim_{k\rightarrow+\infty} \Fout \ForwardPolicy^{k}\right).\\
\end{eqnarray*}
Beware that at each step, the right hand side may is well defined and each step is legal, even if the sum is infinite.
Therefore,  $\mathbf 1_{S^*}\Foutbar  \leq \mathbf 1_{S^*} \Fout$, $\Foutbar(\Init)=\Fout(\Init)$ and $\Foutbar(\Final)=0\leq \Fout(\Final)$. Thus,  $\Foutbar\leq \Fout$ and thus $\overline{F}\leq F$.
In particular, $\overline F$ is a finite measure and $\mathbb E(\tau)<+\infty$.
Furthermore,
$$\overline R  :=  \overline F(\cdot,\Final) = \Fout (\Init) \mathbb E(\delta_{s_{\tau}})  = F(\cdot, \Final)=R;$$  by Proposition \ref{prop:sampling_dist},  we conclude that  $\overline R=R$ and  $\overline F$ is a $R$-flow. 
The edgeflow $\overline F$ is then a flow  with target $F(\cdot,\Final)=R$. 

Finally, since $\overline F$ only depends on $\ForwardPolicy$ and $\Fout(\Init)=\Foutbar(\Init)$, we have $\overline {\overline F}= \overline F$ so we have equality in the computation above and  $\mathbf{1}_{S^*}\lim_{k\rightarrow+\infty} \overline \Fout \ForwardPolicy^{k}=0$.
\end{proof}

\begin{repprop}{prop:decomposition} A edgeflow $F$ has a maximal  0-subflow hence it can be written $$F = F_0+ F_{min},$$
where $F_0$ is a 0-subflow and $F_{min}$ is a minimal edgeflow.
\end{repprop}
\begin{proof} One can apply Zorn's Lemma on the set of 0-subflows with the usual order on measures. The supremum of a totally ordered family of measures bounded by $F$ is still a measure bounded by $F$ and 0-flow property is preserved by suprema. The set is not empty since the null measure is a 0-flow bounded by $F$ as $F$ is non-negative.  There thus exists a (non-unique) maximal 0-subflow.
\end{proof}

\begin{reptheorem}{theo:cv} Assume $S$ is at most countable.  Let $0\leq R$ be a reward function. Let
$$ \mathcal L_\alpha(F) = \mathcal L_{\text{FM}}(F) + \alpha \mathcal R (F) $$
with $\mathcal L_{\text{FM}}$ a strong Flow Matching loss among $R$-edgeflows, $\alpha\geq 0$ and  $\mathcal R$ such that  $\partial_{F_0} \mathcal R>0$ and $\partial_{F_0} \mathcal L_{\text{FM}}\geq 0$ for all 0-subflows.
Let $\alpha_n \rightarrow 0^+$, then any sequence $(F_n)_{n\in \NN}$ of   $\mathcal L_{\alpha_n}$-minimizing $R$-edgeflow   converges toward an
acyclic R-flow.
\end{reptheorem}
\begin{proof}
Assume by contradiction that for some $n\in \NN$ the local minimizer $F=F_n$ is not minimal and use Proposition \ref{prop:decomposition} to decompose $F = F_0 + F_{min}$. We have
 $$0=\partial_{F_0} \mathcal L_\alpha =  \partial_{F_0} \mathcal L_{\text{FM}} + \alpha \partial _{F_0}\mathcal R. $$
But  $\partial_{F_0} \mathcal R >0$ and $\partial _{F_0} \mathcal L_{\text{FM}}\geq 0$. Contradiction.

A sequence of minimizers of $\mathcal L_{\alpha_n}$ is a minimizing sequence for $\mathcal L$ thus converging by strong Flow Matching hypothesis. Then, by the previous point, each $F_n$ is minimal, by Lemma \ref{lem:acyclic_close} so is the limit hence acyclic.
\end{proof}

\begin{theo}\label{theo:graph_dec_generalized} Let $G=(S,E)$ be a connected directed graph and let $R$ be a summable target distribution.  Then $\mathcal F_R$ is a non-empty affine subspace of $\ell^1(E)$ :
$$ \mathcal F_R = F + H^ 1(G), $$ where $F$ is any element of $\mathcal F_R$. In particular, $\mathcal F_0 = H^1(G)$.
\end{theo}
\begin{proof}
In view of Propositions \ref{prop:flowspace}  and \ref{prop:decomposition}, it suffices to show that $\mathcal F_0=H^1(G)$ and that $\mathcal F_R$ is nonempty.
 To begin with $H^1(G)\subset \mathcal F_0$ trivially.  Take some  $F\in \mathcal F_0$ and some increasing family  $(K_n)_{n\in \NN}$ of finite subset of $S^*$ such that $\bigcup_{n\in \NN}K_n = S^*$ and
 $\bigcup_{n\in \NN} E\cap (K_n\times K_n) = E$.

Take some $n\in \NN$, consider the complete graph $G'=(S',E')$ of set of states $S'^*=K_n\cup\{\infty\}$ and $E' = (K_n\cup\{\infty\})^2$
on which we consider the 0-flow $F'$ defined by
\begin{itemize}
\item $\forall e\in K_n\times K_n, F'(e)=F(e)$
\item $\forall s\in K_n, F'(s,\infty)=F(s,S^*\setminus K_n)$
\item $\forall s\in K_n, F'(\infty,s) = F(S^*\setminus K_n,s)$
\item  $F'(\infty,\infty)=0$.
 \end{itemize}
Take some edge $e_1=(s_0,s_1)\subset K_n$ such that $|F'(e)|>|F'|(K_n, \infty)$.
  By Flow Matching property, there exists another ingoing or outgoing edge $e_2=s_1\rightarrow s_2$ or $e_2=s_1\leftarrow s_2 $ such that $F'(e_2)\neq 0$. We continue this process until we reach some already visited state to construct a path $\gamma = (s_p)_{p\in \lsem 0,q\rsem}$ with $0< q$ containing a cycle $\omega$.
If  $\omega\subset K_n$ then we keep the cycle $\omega\subset \gamma$.
Otherwise, we note that
\begin{eqnarray*}\min_{e\in \omega}|F(e)|&\leq&  |F'(s_{p},s_{p+1})|\\
&\leq&  |F|(K_n,S^*\setminus K_n) \\ &<&|F(e_1)|=|F'(e_1)|.\end{eqnarray*}
So that we may construct another path from $e_0$ in the flow $F'' = F'-\lambda_0\mathbf{1}_{\gamma}$ with $\lambda_0=\min_{e\in \gamma}|F_0(e)|$. Iterate the process, since $G'$ is finite and since $F''$ has one less non-zero edge, the process thus stops on a cycle within $K_n$.
We  can thus consider some $F'_1:=F'-\lambda_0\mathbf{1}_{\omega}$ with $\lambda_0=\min_{e\in \omega}|F'(e)|$ and $\omega\subset K_n$. Note that $F'_1$ contains at least one non-zero edge less than $F'$ and $|F'_1|\leq |F'|$. Iterate the construction to build a sequence $(F'_q)_{q\leq r}$ such that $(|F'_q|)_{q\leq r}$ is decreasing.  Since $G'$ is finite, the process stops and we then constructed a family of cyclic flows $(\lambda_i \mathbf{1}_{\omega_i})_{i\in \lsem 0,\lsem r}$ such that  $F' = \sum_{i=0}^r \lambda_i \mathbf{1}_{\omega_i} + F'_ r$. Since we may not extract any more cycle from $F_r'$, it means that $|F_r'|(e)\leq |F_r'|(K_n,\infty)$ for all $e\in K_n\times K_n$.
Since all cycles $(\omega_i)_{i\leq r}$ are within  $K_n$, we deduce a decomposition for the flow $F$ :
$$F =  \sum_{i=0}^r \lambda_i \mathbf{1}_{\omega_i} + F_1$$ $$ \forall  e\in K_n\times K_n,\quad | F_1|(e)\leq |F_1|(K_n,S^*\setminus K_n).$$

We thus can construct a sequence of 0-flows $(F_n)_{n\in \NN}$ such that $F_0 = F$ and for all $n\in \NN$:
$$ |F_{n+1}|\leq |F_n|, \quad F_{n} =  \sum_{i=0}^{r_n} \lambda_i \mathbf{1}_{\omega_{i,n}} + F_{n+1},$$
$$ \forall  e\in K_n\times K_n,\quad | F_{n+1}|(e)\leq |F_{n+1}|(K_n,S^*\setminus K_n).$$

We deduce from those properties that $\forall n\in \NN, |F_n|\leq |F|$ and  $\forall \in E, \lim_{n\rightarrow +\infty} F_n(e) = 0$, therefore $F_n\rightarrow 0$ in $\ell^1(E)$ so that
$$ \exists (\lambda_i)_{i\in \NN}, (\omega_{i})_{i\in \NN},\quad  F= \sum_{i=0}^{\infty} \lambda_i \mathbf 1_{\omega_i},$$
e.g. $F\in H^1(G)$.

To construct an element of $\mathcal F_R$, for each final edge, $e=s \rightarrow \Final$ choose a path $\gamma_e : = \Init \rightarrow \cdots \rightarrow s \rightarrow \Final$ which exists since $G$ is connected. Then define $F= \sum_{e} R(e) \mathbf{1}_{\gamma_e}$. By construction, $F$ is a $R$-flow.

\end{proof}
\begin{repprop}{prop:graph_dec} Let $G=(S,E)$ be a connected directed graph and let $R$ non-negative reward function.  Then $\mathcal F_R^+$ is a  non-empty  convex domain of $\mathcal F_R$.
Furthermore,
$$ \mathcal F^+_R = \mathcal F^+_{R,acyclic} + H^1_+(G).$$
\end{repprop}
\begin{proof}
Convexity follows from convexity of the cone $\{F\in \ell^1(E) ~|~ F\geq 0\}$ and the set $\mathcal F_R$ as $$\mathcal F^+_R = \mathcal F_R \cap \{F\in \ell^1(E) ~|~ F\geq 0\}.$$

The element of $\mathcal F_R$ constructed in the proof of the previous proposition is actually non-negative. To show that $\mathcal F_0^+ = H^1_+(G)$, the same proof works, positivity ensuring we can choose positively weighted simple loops. Minimality implies acyclicity so Proposition \ref{prop:decomposition} gives the wanted results.

\end{proof}
\begin{replemma}{lem:acyclic_close} If $S$ is at most countably infinite, then  the subsets of acyclic edgeflows  and minimal edgeflows are both closed for narrow convergence.
\end{replemma}
\begin{proof}
In view of the preceding Proposition, minimal is equivalent to acyclic  so we only need to prove the acyclic case.
Consider a sequence of edgeflows $F_n$ converging toward  some $F_\infty$. If $F_\infty$ admits a cycle $\gamma = s_0\rightarrow_1\rightarrow \cdots \rightarrow s_p$. Then. For all $i\in \Z/p\Z$, $F_n(s_i,s_{i+1})\xrightarrow{n\rightarrow +\infty} F_{\infty}(s_i,s_{i+1})>0$ there thus exists some $N\in \NN$ such that  $\forall n\geq N, \forall i\in \Z/p\Z, F_n(s_i,s_{i+1})>0$. Therefore, for $n$ big enough $\lambda \mathbf {1}_\gamma \leq F_n$ with $\lambda = \min_{i\in \Z/p\Z} F(a_i,a_{i+1}) $.

\end{proof}

\subsection{Proof of Theorem \ref{theo:weak_diver}}
Recall that $f,g$ are functions $\RR_+\rightarrow \RR_+$ which are zero exactly at 1.

\begin{reptheorem}{theo:weak_diver} 
If $\Div_f$ is a proper divergence, then  $\mathcal L_{\text{FM},f}, \mathcal L_{\text{DB},f}$  are \textbf{unstable} except if $\Div_f$ is the total variation.

The losses $\mathcal L_{\text{FM},g,\nu},\mathcal L_{\text{DB},g,\nu}$ and $\mathcal L_{\text{TB},g,\nu}$ are unstable.

\end{reptheorem}
We prove this Theorem under simple technical, nonminimal assumptions covering reasonably many implementations. Those assumptions are motivated by the simplest use of a Lebesgue convergence Theorem. 
\begin{prop}\label{prop:weak_FM_f_diver} 
If $\Div_f$ is a proper divergence, then the minimizers of  $\mathcal L_{\text{FM},f}$  among edgeflows uch that $\Fin\ll \Fout$ are flows. Furthermore,
\begin{itemize}
 \item assume $f$ is continuous and continuously differentiable piecewise;
 \item assume $\frac{d \Fin}{d \Fout}$ is bounded;
\end{itemize}
  then for any non-trivial 0-subflow $F_0$ of  $F$ the directional derivative $\partial _{F_0} \mathcal L_{\text{FM},f}(F)$ exists and
$$\partial _{F_0} \mathcal L_{\text{FM},f}(F) \leq 0$$
with equality for all such $F, F_0$ and for all transition constraints if and only if $\Div_f$ is proportional to total variation.

Finally, on edgeflows such that $\Fout\ll \nu$ and $\nu \ll \Fout$, $\mathcal L_{\text{FM},g,\nu}$ is unstable.

\end{prop}
\begin{prop}\label{prop:weak_DB_f_diver} 
If $\Div_f$ is a proper divergence,  on flows of the form $F_f := \Fout\otimes \pi_f$ and $F_b := \Fout\otimes \pi_b$ where $\pi_f$ and $\pi_f$ are independent and such that $F_b\ll F_f$ then the minimizers of $\mathcal L_{\text{DB},f}(F_f,F_b)$ are flows. Furthermore,
\begin{itemize}
 \item assume $f$ is continuous and continuously differentiable piecewise;
 \item assume $\frac{d F_b}{d F_f}$ is bounded;
\end{itemize}
  then for any non-trivial 0-subflow $F_0$ of  $F_f$ the directional derivative $\partial _{F_0} \mathcal L_{\text{DB},f}(F_f,F_b)$ exists and
$$\partial _{F_0} \mathcal L_{\text{DB},f}(F_f,F_b) \leq 0$$
with equality for all such $F_f,F_b,F_0$ and for all transition constraints if and only if $\Div_f$ is proportional to total variation.

Finally, on edgeflows such that $\Fout\ll \nu$ and $\nu \ll \Fout$, $\mathcal L_{\text{FM},g,\nu}$ is unstable.

\end{prop}
While the admissibility is simply a measurable translation of arguments one may find in \cite{bengio2021gflownet}, the instability is a consequence of the following two Lemmas. 

\begin{lemma} Let $\mathcal X$ be a non-empty measurable space, let $F_0,F_1,F_2$ be three finite non-negative measures on $\mathcal X$.
If  $\Div_f$ is a proper divergence and 
\begin{itemize}
    \item $f$ is continuous and piecewise continuously differentiable
    \item $ F_2 \ll F_1\geq F_0$
\end{itemize}
then  
$$ \partial_{(F_0,F_0)} \Div_f(F_2,F_1) \leq 0.$$
The equality is realized for $\Div_f=\Div_{TV}$. Furthermore,
if there exists  $F_0,F_1$ as above such that for all $\alpha \in L^1(F_1)$ the equality holds for $(F_0,F_1,F_2=\alpha F_1)$,  then $\Div_f\propto \Div_{TV}$.
\end{lemma}
\begin{proof}

Define $$\phi(t) = \int_{S^*} g\left( \frac{d (F_2+tF_0)}{d(F_1+t F_0)} \right)\mathrm d (F_1+tF_0)$$
and  $F_2= \alpha F_1$ and $F_0 = \beta F_1$.
\begin{eqnarray*}
\phi(t)&=&  \int_{\mathcal X} g\left(\frac{ \alpha+t\beta}{1+t\beta} \right) (1+t\beta)\mathrm d F_1\\
	&=&  \int_{\mathcal X} g\left(1+\frac{ \alpha-1 }{1+t\beta} \right) (1+t\beta)\mathrm d F_1\\
\phi'(t)&=&  -\int_{\mathcal X} g'\left(1+\frac{ \alpha-1 }{1+t\beta}  \right)\frac{ (\alpha-1) (1+t\beta) }{(1+t\beta)^2}\beta\mathrm d F_1\\
&&\quad + \int_{\mathcal X} g\left(1+\frac{ \alpha-1 }{1+t\beta}  \right)\beta\mathrm d F_1\\
&=& \int_{\mathcal X} \psi\left(\frac{ \alpha-1 }{1+t\beta}  \right) \mathrm d F_0,
\end{eqnarray*}
where $\psi(x) = g(1+x)-xg'(1+x)$.
Note that $$t,x\mapsto g\left(\frac{ \alpha(x)+t\beta(x)}{1+t\beta(x)} \right) (1(x)+t\beta(x))$$
is measurable in the variable $x$ and continuous in the variable $t$ except at isolated point $t$ and that its derivative with respect to $t$ is bounded in a neighborhood of $t=0$ since $\alpha$ is bounded by hypothesis and $\sup \beta \leq 1$ by construction. Therefore the derivation under the integral  sign is legal.

Since $g$ is convex, $\psi$ is non-decreasing on $\RR_-$ and non-increasing on $\RR_+$ hence $\psi\leq \psi(0)=0$. Therefore, $\phi'(t)\leq 0$ so that $\partial _{F_0,F_0} \Div_f(F_2,F_1) \leq 0$.

Let $0\neq F_0\leq F_1$  such that for all $\alpha\in L^1(F_1)$ equality holds for $(F_0,F_1,F_2=\alpha F_1)$.  Take $\alpha_C = C \mathbf 1_{\mathcal X}$  for $C\in \RR_+$. Since equality holds for $(F_0,F_1,\alpha_C F_1)$ then  for all $C\geq 0$ we have  
$$0=\phi'(0) = \int_{\mathcal X} \psi\left(C-1 \right) d F_0$$
thus $\forall x>-1, \psi(x)=0$. 
 Under our regularity hypothesis this implies that $g$ solves the differential equation $y+(x-1)y'=0$, we deduce that $\forall x\in \RR_+^*,  g(x) = \omega(x) |1-x|$  with $\omega$ piecewise constant. It follows from continuity that $\omega$ may only be discontinuous at $1$ so we may rewrite $g$ as  $g(x) = \omega_1 |1-x| + \omega_2 (1-x)$ hence the result.

\end{proof}
\begin{lemma} Let $\mathcal X$ be a non-empty measureable space.
Assume that
\begin{itemize}
    \item $g$ is continuous and piecewise continuously differentiable;
    \item $g\geq 0$ with equality only at $1$;
\end{itemize}
then for any $F_0,F_1$ with $0 \neq F_0\leq F_1$, there exists $ F_2 \ll F_1$ such that
$$ \partial_{(F_0,F_0)} \Div_{g,\nu}(F_2,F_1) < 0.$$
\end{lemma}
\begin{proof}
The same computation as in the preceding Lemma yields $\partial_{F_0,F_0} \Div_{g,\nu}(F_2,F_1)= \int_{\mathcal X} \psi\left(\frac{\alpha-1}{1+t\beta}\right)d F_0$ with $\psi(x) = -xg'(1+x)$. Again, if $ \partial_{(F_0,F_0)} \Div_{g,\nu}(F_2,F_1)\geq 0$ for all $F_2\ll F_1$ then $\psi\geq 0$ hence $g$ admits its maximum at $1$. 
Finally, $0\leq g \leq g(1) = 0$ so $g$ is zero everywhere, contradicting our hypotheses.
\end{proof}

We do not have such strong generic instability results for TB loss as we do for DB and FM losses. This is in part due to the nonlinearity of the operator $F\mapsto F^\otimes$ that associates the trajectory flow to a given edgeflow.

\begin{lemma} Let  $F_f,F_b$ be edgeflows on some $S$ with transition constraints $\mu$ and let $F_0$ be 0-subflow such thats $\mathbf 1_{S^*\times S^*}\mu\ll F_0$.
Assume that 
\begin{itemize}
    \item $F_f\ll F_b$ and  $\frac{d F_f}{d F_b}$ is bounded,
    \item for all $\xi\in \RR_+$,
    $$\int_{\underline s\in \mathrm{Path}(\Init,\Final)} g\left(e^{\xi\mathrm{len}(\underline s)}\right) d\nu(\underline s)<+\infty,$$
\end{itemize}
 then: 
\begin{multline*}
	\lim_{c\rightarrow +\infty}\mathcal L_{\text{TB},g,\nu}( (F_f+c F_0)^{\otimes}, (F_b+c F_0)^{\otimes}) = \\ 
	\int_{s\in S^*\times S^* }g\left[\frac{d F_f}{d F_b }(s_{0}\rightarrow s)\frac{d F_f}{d F_b }(s'\rightarrow s_f)\right] d\omega(s,s')
	\end{multline*}
with $\omega(\cdot) = \phi\# \nu$ and  is the map $\phi(s_0\rightarrow s_1\rightarrow \cdots\rightarrow s_f) = (s_1,s_\tau)$.

In particular, if $F_f(s_0\rightarrow \cdot ) = F_b(s_0\rightarrow \cdot)$ and $F_f(\cdot \rightarrow s_f ) = F_b(\cdot\rightarrow s_f)$ then the limit above is 0.
\end{lemma}
\begin{proof}
   A direct computation yields:
\begin{multline*} \frac{d F_f^\otimes }{dF_b^\otimes}(s_0\rightarrow \cdots s_\tau\rightarrow \Final) = \\  \prod_{t=1}^{\tau+1}\left(\frac{d F_f}{dF_b}(s_{t-1}\rightarrow s_t)\right) \prod_{t=1}^{\tau}\left(\frac{d F_{b,\mathrm{in}}}{dF_{f,\mathrm{out}}}(s_t)\right).
\end{multline*}
Write  $P:=\mathrm{Path}(\Init,\Final)$ and paths $s_0\rightarrow \cdots\rightarrow s_\tau\rightarrow s_f$ by $\underline s$. Then, writing 
\begin{multline*}
    A_c(\underline s) :=   \prod_{t=1}^{\tau+1}\left(\frac{d (F_f+cF_0)}{d(F_b+cF_0)}(s_{t-1}\rightarrow s_t)\right) \\ 
    \times \prod_{t=1}^{\tau}\left(\frac{d (F_{b,\mathrm{in}}+cF_0) }{d(F_{f,\mathrm{out}}+cF_0) }(s_t)\right)
\end{multline*}
and 
$$ B(\underline s) := \frac{d F_f }{d F_b}(s_{0}\rightarrow s_1)\frac{d F_f}{d F_b}(s_\tau\rightarrow s_f),$$
we have 
\begin{multline*} \int_{\underline s \in P}g\left[\frac{d (F_f+cF_0)^\otimes }{d(F_b+cF_0)^\otimes}(s_0\rightarrow \cdots s_\tau\rightarrow \Final)\right]d\nu(\underline s)  \\ 
= \int_{\underline s \in P}g\left[B(\underline s)\right]d \nu(\underline s)  \\
+\int_{\underline s \in P}\left\{g\left[A_c(\underline s)\right] - g\left[B(\underline s)\right]\right\}d \nu(\underline s).
\end{multline*}
On the one hand,  $B(s_0\rightarrow s_1\rightarrow \cdots \rightarrow s_\tau \rightarrow s_f)$ only depends on $s_1$ and $s_\tau$, therefore
\begin{multline*}
\int_{\underline s \in P}g\left[B(\underline s)\right]d \nu(\underline s) = \\\int_{\underline s \in P}g\left[\frac{d F_f}{d F_b }(s_{0}\rightarrow s)\frac{d F_f}{d F_b }(s'\rightarrow s_f)\right]d \omega(s,s').    
\end{multline*}
On the other hand, since $F_0$ is a 0-flow 
in particular $F_0(s_0\rightarrow \cdot) =F_0(\cdot \rightarrow s_f)=0$ but since $\textbf 1_{S^*\times S^*}\mu \ll F_0$, then 
for $\mu$-almost all $s\rightarrow s'$ from $S^*$ to $S^*$, 
$\frac{d( F_f+cF_0)}{d(F_b +cF_0)} (s\rightarrow s')=\frac{a+c\delta}{b+c\delta}$ for some $a,b\geq 0$ and $\delta>0$. 
Therefore all terms  in 
$A_c(\underline s)$ tend towards 1 except $\frac{d F_f}{dF_b}(s_{t-1}\rightarrow s_t)$ and $\frac{d F_f}{dF_b}(s_\tau\rightarrow s_f)$;
and thus $$\forall \underline s\in P, \quad \lim_{c\rightarrow +\infty} A_c(\underline s) = B(\underline s).$$
Since we assumed $\frac{d F_f}{d F_b}$ bounded, it follows that $\left|g[A_c(\underline s)]-g[B(\underline s)]\right|\leq  C g\left(e^{\xi \mathrm{len}(\underline s)}\right)$ for some $C,\xi>0$ independent from $\underline s$. The result then follows by 
Lebesgue convergence Theorem.
\end{proof}

This Lemma implies instability of the loss $\mathcal L_{\text{TB},g,\nu}$ as this limit may be bigger or smaller than the value of the loss for $c=0$ depending on the relative values of $F_f$ and $F_b$.  Depending on the actual implementation, one may be able to enforce the initial/terminal flows constraints $F_f(s_0\rightarrow \cdot ) = F_b(s_0\rightarrow \cdot)$ and $F_f(\cdot \rightarrow s_f ) = F_b(\cdot\rightarrow s_f)$. This is the case in our implementation of the TB loss on hypergrids.

\subsection{Proof of Theorem \ref{theo:stable_AB}}
We recall the assumptions in our family of putative stable losses.
\begin{itemize}
    \item $R\ll\nu$ so that $\nu$ has ``full support" with respect to $R$ and $\frac{dR}{d\nu}$ is bounded,
    \item $f\geq 0$ and $g\geq 1$,
    \item $f(x)=0\Leftrightarrow x=0$;
    \item $f$ and $g$ are continuous and piecewise continuously differentiable;
    \item $f$ decreases on $\RR_-$ and increases on $\RR_+$ 
    \item $\partial_{(1,1)}g \geq 0$
\end{itemize}

\begin{reptheorem}{theo:stable_AB} If $f,g,\nu,R$ satisfy the properties above then  for $R$-edgeflows the loss $\mathcal L_{\text{FM},\Delta, f, g},\mathcal L_{\text{DB},\Delta, f, g}$ are stable.
\end{reptheorem}

This theorem is a direct consequence of the following Lemmas.
\begin{lemma} Let $\mathcal X$ be a measurable space, the function $\alpha,\beta \mapsto \int_{x\in \mathcal X} f(\alpha(x)-\beta(x)) g(\alpha(x),\beta(x)) d\nu(x)$ defined on $L^\infty(\mathcal X,\nu)\times L^\infty(\mathcal X,\nu)$ is distance-like. 
\end{lemma}
\begin{proof}
Let $\alpha,\beta\in L^{\infty}(\nu)$, since $f,g$ are non-negative then so is $I:=\int_{x\in \mathcal X} f(\alpha(x)-\beta(x)) g(\alpha(x),\beta(x)) d\nu(x)$. Assuming that $I=0$ we have $f(\alpha(x)-\beta(x))g(\alpha(x),\beta(x))=0$ for $\nu$-almost all $x$. Since $g\geq 1$ and $f(y)=0 \Leftrightarrow y=0$, we conclude that $\alpha(x)=\beta(x)$ for $\nu$-almost all $x$ hence $\alpha=\beta$.
\end{proof}
\begin{lemma}
     Let $\mathcal X$ be a measurable space, for any $\alpha,\beta,\gamma\in L^\infty(\nu)$ non-negative,
     the function $\phi:t \mapsto \Delta_{f,g,\nu}(\alpha+t\gamma,\beta+t\gamma)$ has non-negative derivative at $t=0$.
\end{lemma}
\begin{proof}
Apply the Lebesgue derivation Theorem under the integral sign.
\end{proof}

\section{Experiment Details}
\subsection{Hypergrid}

\subsubsection{Implementation Details}
GFlowNets for such small graphs were implemented using the adjacency matrix of the underlying graph. The flow itself is implemented as an array containing the weight of each transition. Implementation is done with v2.4.4 of tensorflow  \cite{tensorflow2015-whitepaper}. Minimization was done using Adam with a learning rate 0.01 and default parameters otherwise.
Various reward distributions have been tested with little impact on the end result.  Self-training is emulated by updating on regular interval the training distribution $\nu$ to an ``optimistic'' reachable part density of the flow e.g. $\mu = \overline {\Fout+\delta}$ with $\delta>0$, we used $\delta=0.001$. We did not try to optimize  $\delta$, and it is probable that its optimal value depends on the choice of the Loss.  Each epoch updates $\mu$  if self-training is switched on, then proceed with 200 gradient steps. 

\subsubsection{Metrics}

We compared Flow Matching losses $\log\mathcal L$, sampling distribution error 
$\mathcal E_F := \log\sum_{s\in S^*}\left|\mathbb P(M_\tau = s)-\frac{R(s)}{R(S^*)}\right| = 
 \log \sum_{s\in S^*}\left| \frac{\overline R(s)}{\overline R(S^*)}-\frac{R(s)}{R(S^*)}\right |$, 
 the averaged sampling time  $\mathbb E(\tau)= \frac{\overline \Fout(S^*)}{\overline R(S^*)}$, the reward error $\mathcal E_{ R} := \sum_{s\in S^*}\frac{\left| \widehat R(s) - R(s) \right|}{R(S^*)} $
 and the initial flow error $\mathcal E_{I}=\log\left |  \frac{F(\Init\rightarrow S^*)-R(S^*)}{R(S^*)} \right|$.
Here $\widehat R = \max(F(S\rightarrow \cdot) - F(\cdot\rightarrow S^*),0)$, which is thought of as an estimation of the reward at a given state. 
To estimate $\mathcal E_F$ and  $\mathbb E(\tau)$  one has to compute $\overline R$, which is deduced from $\overline{ \Fout}$ and the Markov chain matrix $\ForwardPolicy$. The former is could be computed using R ${F_{out}}_{|S^*} = F(\Init \rightarrow \cdot)_{|S^*} (Id_{S^*}- {\ForwardPolicy}_{|S^*})^{-1}$ but  this becomes rapidly too computationaly expensive. Instead, we use a power method to approximate $\overline \Fout$ with a cutoff at $\lambda W$ and  $\lambda >1$ with the rationale that a good sampler should not need much more step than the width. We usually take $\lambda \in \{4,10\}$ to give ourselves some margin.

 Although some of those estimators are redundant, their interpretation is different:
 \begin{itemize}
     \item $\mathcal E_F$ measures how good  the flow is as a sampler;
 \item $\mathcal E_{ R}$ measures how good  the flow is as an approximator;
  \item $\mathcal E_{I}$ measures how good  the flow is as an integrator;
 \item $\mathbb E(\tau)$ measures how efficient the flow is; $\mathcal L$ measures how far from completion the training is. 
 \end{itemize}

\subsection{Cayley Graphs}
\subsubsection{Implementation Details}
The implementation chooses an injective group morphism $\rho : \mathfrak S_p \rightarrow \mathrm{GL}(\RR^d)$ for some $d$ and a point $P\in \RR^d$ so that each $g\in S^*$ is associated to $\rho(g)P$. 
The trainable edgeflow $F$ is then a function $F^\theta : \RR^d \rightarrow \RR_+^q$ so that $F^\theta (\rho(g)P) = \left(F(g\rightarrow \sigma_i g)\right)_{i\in \lsem 1,q\rsem}$. The ingoing and outgoing flow are tractable for small $q$. In order to train the flow, a batch of paths are generated using the forward policy the flow neglecting transition to the sink $s_f$ so that all paths reach a cutoff length. The states reached by such an unstopped path $(\underline s_t)_{t\geq0}$ are weighted by the probability that $t\leq \tau$ conditioned by ``$\forall t\leq \tau, s_t=\underline s_{t}$". This allows to work with fixed size state batchs and reduce the variance of the estimator of a loss with $\nu$ given by the density of paths generated by the flow itself.

The experiments were conducted using  $\rho(\sigma):= (\sigma(1),\cdots,\sigma(p))$ and $F$ given by a dense MLP of depth 3 and width 32 with LeackyReLU activations,  training rate 1e-2 and path batch size 64.  

We compared $\LossFM$ to $\mathcal L_{\text{FM},\Delta, f, g}$ for a range for choices of function $f,g$ of the same type as for hypergrids. 
The initial flow is fixed up to a scalar parameter $F(s_0\rightarrow \cdot)=\Finit \mathcal U(S^*)$ where $\mathcal U(S^*)$ is the uniform distribution on $S^*$. This allows to have better exploration and emulate the resolution of a generic problem (eg finding a path to reward starting from any position).

\subsubsection{Comparison of Stable Losses}
Multiple variations of the stable loss introduced above were tested. We observed that tuning the hyperparameters was challenging, high stabilizing regularization was useful. The choice of power for the regularization part $g$ in the result given above on Cayley graphs has a drawback: since the power is smaller than one, small flows tend to implode to 0. 
Interestingly, the flow still has good sampling properties as the above shows. Non-imploding stable losses ($g(x)=1+|x+y|^\beta$ with $\beta=0$ or $\beta>1$) showed non-exploding behavior but suboptimal expected reward.






\end{document}